# UDHF$^2$-Net: Uncertainty-diffusion-model-based High-Frequency TransFormer Network for Remotely Sensed Imagery Interpretation

Pengfei Zhang [a], Chang Li [a] *, Yongjun Zhang [b], Rongjun Qin [c], Kyle Gao [d] and Jonathan Li [d]

[a] *Key Laboratory for Geographical Process Analysis & Simulation of Hubei Province, and College of Urban and Environmental Science, Central China Normal University, Wuhan, China;*

[b] *School of Remote Sensing and Information Engineering, Wuhan University, China.*

[c] *Department of Civil, Environmental and Geodetic Engineering; Department of Electrical and Computer Engineering; Translational Data Anayltics Institute, The Ohio State University, Columbus, Ohio, United States of America*

[d] *Department of Systems Design Engineering; University of Waterloo, Waterloo, Canada*

** Corresponding author: Chang Li, lcshaka@126.com && lichang@ccnu.edu.cn*

**Abstract:**

Remotely sensed imagery interpretation (RSII), including semantic segmentation and change detection, faces the three major problems: (1) objective representation of spatial distribution patterns with coexistence of spatial stationary and non-stationary; (2) edge uncertainty problem caused by downsampling encoder and intrinsic edge noises (e.g., mixed pixel and edge occlusion etc.); and (3) false detection problem caused by geometric registration error in change detection. To solve the aforementioned problems, uncertainty-diffusion-model-based high-Frequency TransFormer network (UDHF$^2$-Net) is the first to be proposed, whose superiorities are as follows: (1) a spatially-stationary-and-non-stationary high-frequency connection paradigm (SHCP) is proposed to enhance the interaction of spatially frequency-wise stationary and non-stationary features to yield high-fidelity edge extraction result. Inspired by HRFormer,

SHCP proposes high-frequency-wise stream to replace high-resolution-wise stream in HRFormer through the whole encoder-decoder process with parallel frequency-wise high-to-low streams, so it improves the edge extraction accuracy by continuously remaining high-frequency information; (2) a mask-and-geo-knowledge-based uncertainty diffusion module (MUDM), which is a self-supervised learning strategy, is proposed to improve the edge accuracy of extraction and change detection by gradually removing the simulated spectrum noises based on geo-knowledge and the generated diffused spectrum noises; (3) a frequency-wise semi-pseudo-Siamese UDHF$^2$-Net is the first to be proposed to balance accuracy and complexity for change detection. Besides the aforementioned spectrum noises in semantic segmentation, MUDM is also a self-supervised learning strategy to effectively reduce the edge false change detection from the generated imagery with geometric registration error. In semantic segmentation experiments, UDHF$^2$-Net achieves the best mIoU values of 92.00%, 87.31% and 55.27% respectively on ISPRS Potsdam, Vaihingen and LoveDA datasets compared with SOTA networks (e.g., SkySense, FT-UNetFormer and HRFormer etc.). In change detection experiments, semi-pseudo-Siamese UDHF$^2$-Net also achieves the best IoU values of 93.12%, 90.47% and 78.13% respectively on WB-CD, LEVIR-CD and SYSU-CD datasets compared with SOTA networks (e.g., P2V-CD, ChangeFormer, and ChangeCLIP etc.). In both semantic segmentation and change detection ablation experiments, UDHF$^2$-Net achieves the gains at least by 1.07% in mIoU and 1.23% in IoU than the network without MUDM respectively, and at least 0.94% in mIoU and 1.24% in IoU than the single-branch spatially stationary or non-stationary high-frequency connection networks respectively. Moreover, it increases 2.35% mIoU than using differencing network architecture for change detection ablation experiments. This study could contribute to enriching the theory and methods of frequency-domain-based high-accuracy RSII.

## 1. Introduction

Remotely sensed imagery interpretation (RSII) is a significant foundational research to further realize highly-accurate planetary observation, which focuses on extracting both static and dynamic surface information from remotely sensed imagery with computer technology (Mallet and Chehata, 2024; Weng, 2018). RSII generally includes the following primary tasks: semantic segmentation and change detection. The former is a pixel-wise classification task to conduct static image interpretation with a single temporal image (Ma et al., 2019). And change detection pays more attention to whether the pixels of bi-temporal images have changed to perform dynamic image interpretation (Shi et al., 2020). There are two widely used approaches for change detection. One is the post-classification approach, which compares the classification results of bi-temporal images to yield change mapping (Wu et al., 2017); The other approach is image differencing, which classifies the differencing result of bi-temporal images to generate change or no-change categories (Zhang et al., 2020). The aforementioned approaches are regarded as binary classification issue. Therefore, semantic segmentation and change detection are essentially the same task, which share a common goal: pixel-wise classification.

With the rapid advancement of modern machine learning, deep neural networks (DNNs) have emerged as a cutting-edge and hot-topic technology in the field of RSII due to their inherent strengths of capturing high dimensional abstract features (Wang et al., 2024a). Currently, many scholars are dedicated to explore the research of RSII with DNNs from several perspectives:

(1) Network architecture design. Encoder-decoder is a common architecture to perform direct classification for semantic segmentation and change detection, which could generate multi-scale features to capture widely contextual and spatial information for improving the sense understanding ability (Chen et al., 2018). Meanwhile, numerous scholars have adopted double-branch architecture comprising two extended encoders as a variant of encoder-decoder to conduct change detection, which has demonstrated remarkable efficacy in characterizing change features by post-classification or differencing methods (Tian et al., 2022).

(2) Feature domain selection. DNNs have become the mainstream approach of RSII and can be classified into the following types: spatial-wise and frequency-wise DNNs. The former relies on convolutional neural networks (CNNs) and Transformer networks to extract local and global spatial-wise features (Cheng et al., 2024). Conversely, the later depend on periodic signals or non-periodic signals to achieve spatially frequency-wise stationary or non-stationary features. Compared with spatial-wise DNNs, frequency-wise DNNs have shown powerful potentiality to perform RSII due to their representation ability of rich edge and structural information (Chen et al., 2024).

The aforementioned researches have provided us with valuable insights from various perspectives. However, there still are several key details of RSII shown in Fig.1 that are worth deeper exploring as follows:

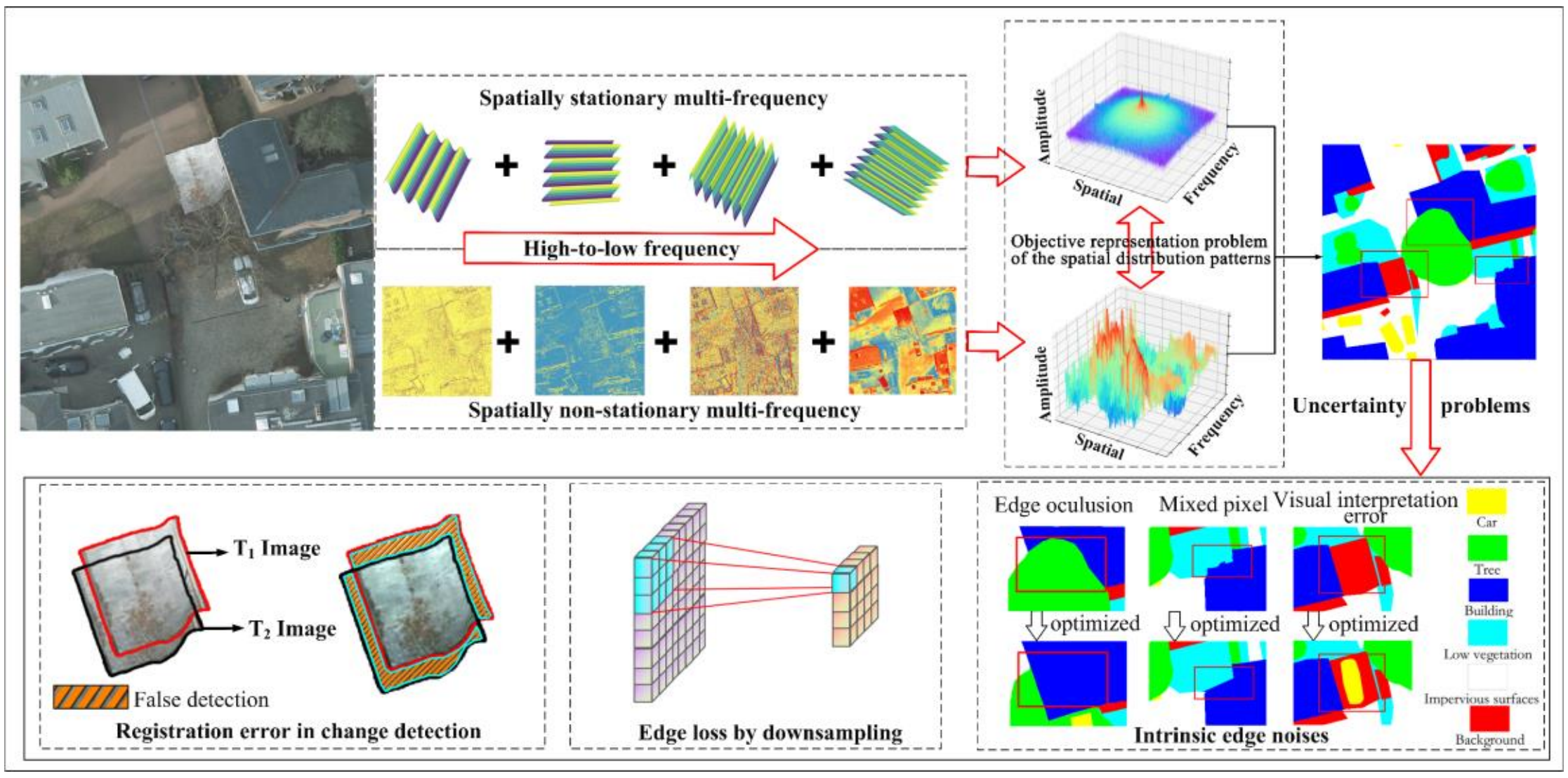

**Figure 1.** The main problems of RSII. (1) objective representation problem of the spatial distribution patterns with the coexistence of spatial stationary and non-stationary; (2) edge uncertainty problem caused by downsampling encoder and radiometric noises of intrinsic edge (i.e., mixed pixels, edge occlusion and visual interpretation errors); and (3) false detection problem caused by geometric registration error in change detection.

(1) How to objectively represent the spatial distribution patterns with the coexistence of spatial stationary and non-stationary? Compared with single frequency-wise feature representation, the coexistence of spatial stationary and non-stationary is more in line with objective laws. However, there are no existing studies that effectively integrate spatially frequency-wise stationary and non-stationary features. Spatially frequency-wise stationary features are essential for representing global consistent information, such as structure, while non-stationary features could capture local abrupt changes, such as edges. Therefore, only single spatially frequency-wise feature couldn't fully describe the completeness of remote sensing scenes. Current researches that focus exclusively on either spatial frequency-wise stationary or non-stationary information restrict the ability of completely characterizing complex remote sensing scenes.

(2) How to reduce the loss of edge details due to downsampling operations of DNNs with classical encoder-decoder architecture? Downsampling operations embedded in encoder, such as max or average pooling, strided or dilated convolutions, are commonly leveraged to reduce the resolution of feature maps while capturing broader contextual information. However, this way makes it less effective at retaining sharp boundaries and critical fine details. To address this shortcoming, Wang et al. (2020) and Yuan et al. (2021) proposed HRNet and HRFormer to adopt the high-resolution representation to successfully obtain lossless high-resolution features. However, the loss of edge details still isn't considered to furthermore improve.

(3) How to alleviate radiometric noises of intrinsic edge caused by the following geo-knowledge: mixed pixel, edge occlusion and visual interpretation errors? Radiometric noises in edge increase the identification difficulty for ground objects due to the incorrect spectrum representation. Mixed pixels generally result in blurred boundaries in the transitional zones between different categories and increase classification uncertainty (Hong et al., 2021); Occlusion could cause false positives or negatives (Li et al., 2023). Visual interpretation errors introduce inaccuracies in training data labeling, leading to biased model training and decreased edge extraction accuracy (Lyu et al., 2020). Hence, scientifically alleviating the impact of these radiometric noises is still worth significant efforts to explore for numerous scholars.

(4) How to address the edge false changes caused by misregistration between bi-temporal images? Due to the impact of sensor angles and atmospheric conditions, etc., geometric registration error could arise, which leads to the false changes especially in edge regions. It’s easily leads to higher false detection rates and reduces accuracy of change detection. Besides, Siamese network is a popular paradigm, and easily overlooks the unique spectrum characteristic of bi-temporal images with shared weights (Shen et al., 2022). In contrast, pseudo-Siamese networks could increase computational costs with no-shared weights. Therefore, balancing feature differences and computational efficiency remains a point worth considering.

To overcome the aforementioned problems, it is a complex and systematic task, so an uncertainty-diffusion-model-based high-Frequency TransFormer network (UDHF$^2$-Net) is the first to be proposed to systematically improve each detail of RSII tasks (i.e., semantic segmentation and change detection). The innovations and contributions are listed as follows:

(1) To objectively represent the spatial distribution patterns and address the edge detail loss problem, we propose a spatially-stationary-and-non-stationary high-frequency connection paradigm (SHCP) to enhance the interaction of spatially frequency-wise stationary and non-stationary features to yield high-fidelity edge extraction result. The proposed paradigm has the following advantages:

Firstly, edge details are preserved and extracted by continuously remaining high-frequency information through the whole encoder-decoder process. Especially, SHCP adopts the continuous high-frequency-wise stream to replace the high-resolution-wise stream, and adopts the interaction of multiple parallel high-to-low frequency-wise streams to achieve high-fidelity edge details and reduce edge loss by downsampling operations;

Secondly, natural scenes are objectively represented by fusing spatially frequency-wise stationary and non-stationary features. Compared with the limitations of the single spatially frequency-wise stationary or non-stationary features, our approach objectively represents global and local information by jointing spatially frequency-wise stationary and non-stationary features to objectively and comprehensively describe complex remote sensing scenes.

(2) To address radiometric noises of intrinsic edge, we propose a mask-and-geo-knowledge-based uncertainty diffusion module (MUDM) to gradually remove the simulated spectrum noises based on geo-knowledge and the generated diffused spectrum noises for improving the accuracy of semantic segmentation and change detection. This module has the following advantages:

Firstly, edge uncertainty is reduced by the proposed self-supervised learning strategy. Based on geo-knowledge, several radiometric noises of intrinsic edge (i.e., mixed pixel, edge occlusion and visual interpretation error) and their labels are generated to train and enhance the anit-noise capability.

Secondly, the edge extraction accuracy is improved by the proposed mask-and-geo-knowledge-based uncertainty diffusion module (MUDM) by recovering truth label from noise in the uncertain regions generated based on diffusion.

(3) To address the false change detection problem, a semi-pseudo-Siamese UDHF$^2$-Net with a self-supervised learning strategy is the first to be proposed to detect change in frequency domain, which reduces the false change detection of edge by training the generated imagery with geometric registration error and the aforementioned spectrum noises. It adopts double-branch architecture to expand the ability of high-fidelity edge extraction in semantic segmentation to perform change detection, which also has the following advantages:

Firstly, model complexity is reduced and features of bi-temporal image are effectively extracted by the proposed semi-pseudo-Siamese architecture. The semi-pseudo-Siamese stragery remains the last two layers without shared parameters to train independently.

Secondly, false change detection is adaptively reduced by the proposed MUDM to gradually denoise by the self-supervised learning strategy, i.e., automatically and randomly generated radiometric noises of intrinsic edge, geometric registration noises and diffusion noises of edge for UDHF$^2$-Net training.

**2. Related work**

This section presents a comprehensive review of the literature on DNNs applied to non-dynamic and dynamic RSII, with a particular emphasis on semantic segmentation and change detection tasks.

### *2.1 Semantic segmentation networks*

Semantic segmentation networks aim at classifying each pixel of an image into a specific category, which are usually divided into spatial-wise or frequency-wise DNNs. Because our work pays more attention to exploring frequency-wise feature representation, numerous works on spatial-wise DNNs, which could be used as the comparison networks in the section of experiments and discussion, aren't reviewed.

Many studies have exploited frequency-wise DNNs as following: (1) Spatially frequency-wise stationary feature extraction. Some studies could provide the significant invariant statistical properties that vary with space to capture global features. That ensures a coherent interpretation of unchanging features and is beneficial to capturing global spectrum information, such as Fourier transform, fast Fourier transform. However, it generally neglects the sensitivity in detecting edges within images that undergo significant and abrupt changes; (2) Spatially frequency-wise non-stationary feature extraction. Numerous researches are devoted to obtaining statistical properties of significant changes across different regions for capturing local features, such as short-time Fourier transform, and discrete wavelet transform, etc. That is beneficial to detecting concentrated or localized features such as edges, textures, and patterns that change over space. However, the approach also encounters challenges in maintaining a coherent global representation. In addition, both spatially frequency-wise stationary and non-stationary features contain multiple high-to-low frequency-wise components. Importantly, high-frequency information carries crucial details related to edges and boundaries, which is beneficial to effectively improving edge extraction accuracy (Shan et al., 2021).

In summary, it’s important to fuse the spatially frequency-wise stationary and non-stationary features for objectively represent the spatial distribution patterns of remote sensing scenes. However, no research has been reported that integrates the above advantages and tackles the aforementioned gaps at the same time. Therefore, the joint extraction of spatially frequency-wise stationarity and non-stationarity features offers a new perspective for constructing an advancing and comprehensive framework to harmonize their complementary strengths.

### *2.2 Change detection networks*

Change detection networks strive to assign the binary change labels to the paired pixels of bi-temporal images (Cao and Huang, 2023; Wu et al., 2023). Existing research generally rely on spatial-wise DNNs to perform feature representation and mainly adopts the following architectures to extract features of bi-temporal images:

#### *2.2.1 Single-branch architecture*

Considerable efforts have been dedicated to adopt this architecture to stack bi-temporal images as the input of DNNs by concatenation or image difference for directly performing classification. However, these approaches could introduce geometric noise owning to registration error of bi-temporal images. The insufficiency image registration could result in change error accumulation and further reduce the accuracy of change detection.

#### *2.2.2 Double-branch architecture*

Substantial works have been explored to adopt this architecture by two branches for extracting deep features from bi-temporal images separately, which could be categorized into two types:

(1) Post-classification networks. These networks typically employ two classifiers to generate two classification maps of bi-temporal images, which are then subsequently to obtain the change maps (Shi et al., 2020). It should be noted that the accuracy of change results is directly determined by the effectiveness of the classifier and the image registration error;

(2) Siamese networks. It could be divided into three types based on whether the weights of the two branches are shared: pseudo-Siamese networks with nonshared weights, pure-Siamese networks with shared weights and semi-pseudo Siamese networks. Pseudo Siamese network has more flexibility but has the more computational consumption. And pure-Siamese network could reduce the number of trained parameters and computational complexity, which overlooks the radiometric differences of bi-temporal images. Different from the above networks, semi-pseudo Siamese network is a balance strategy, which adopts partially shared weights of the previous layers and could not share the weights of the last two or three layers to be trained independently. That could allow the networks to adaptively reduce the impact of registration error and maintain fewer computational parameters.

Summarily, existing change detection networks often emphasize on extracting change features in spatial domain. However, there is a substantial untapped potential in exploring frequency-wise change detection networks. Therefore, frequency-wise semi-pseudo-Siamese network could be adopted to adaptively optimize the registration error and learn frequency-wise feature representation for capturing changed information. To our best knowledge, there are currently no report on the frequency-wise change detection DNNs.

### *2.3 Denoising diffusion probabilistic model*

Whether in semantic segmentation or change detection tasks, there are several inherent radiometric noises in edge as follows: (1) mixed pixel in transition zones easily leads to edge location uncertainty of different objects due to limited resolution (Hong et al., 2021), such as farmland and vegetation, grassland and forest land; (2) edge occlusion generally results in semantic uncertainty (Li et al., 2023). For example, local buildings could be identified as vegetation due to occlusion; (3) visual interpretation errors could make different visual interpretation experts to incorrectly label the position and semantic information of the samples, which increase the fitting difficulty of DNNs (Lyu et al., 2020).

Importantly, many scholars have successfully adopted the denoising diffusion probabilistic model to denoise in the field of remote sensing owing to its outstanding denoising performance and robustness, such as image generation (Lu et al., 2024). The overarching goal of a diffusion probabilistic model is to generate a Markov chain for achieving random noise into the input data, and then training this Markov chain to employ variational inference to exhibit exceptional performance (Croitoru et al., 2023).

It consists of two fundamental stages: the forward diffusion process and the reverse denoising process (Ho et al., 2020). During the forward stage, Gaussian noise is continuously added over multiple time steps for training; In the reverse stage, the original data could be reconstructed by minimizing the discrepancy between the discrepancy between the predicted and added noise gradually. Existing approaches aim to generate global noise but lack attention to local uncertain regions, especially edge regions.

To sum up, we propose UDHF$^2$-Net to effectively perform RSII which could be applied in semantic segmentation and change detection tasks. Our superior solution is demonstrated in the following aspects:

(1) Different from current researches of DNNs depending on single types of frequency-wise features, we are the first to combine spatially frequency-wise stationary and non-stationary features for completely describing the remote sensing scenes. Notably, high-frequency features are further retained to deliver lossless edge information for avoiding edge loss due to downsampling encoder and improving the accuracy of edge extraction;

(2) Different from existing researches of DNNs overlooking the impact of intrinsic noises of object edge, we are the first to take fully advantages of denoising diffusion probabilistic model with a self-supervised learning strategy to train generated negative sample including image and lable with simulated intrinsic radiometric noises for enhancing the anti-noise capability and recover the labels for improving extraction accuracy of edges in RSII tasks;

(3) Different from current researches of DNNs not considering the impact of image misregistration, we are the first to adopt the self-supervised learning strategy with denoising diffusion probabilistic model to enhance the anti-false-change-detection capability, and adopt a semi-pseudo-Siamese architecture to considering computational complexity and accuracy improvement of change detection.

## 3. Methodology

The proposed UDHF$^2$-Net is an end-to-end cascade architecture to systematically optimize the process of RSII in each step including remote sensing semantic segmentation and change detection tasks, the overview of which is shown in Fig.2. It contains of the following main components:

(1) Spatially-stationary-and-non-stationary high-frequency connection paradigm (SHCP) is proposed to generate the initial segmentation map. The input image $\mathbf{X} \in \mathbb{R}^{H\times W\times 3}$ could be decomposed into different high-to-low frequency-wise spatially-stationary-and-non-stationary components (i.e., $\mathbf{F}_{\text{stationary}}$ and $\mathbf{F}_{\text{non-stationary}}$) by spatially-stationary-and-non-stationary multi-frequency generation module (SMGM). Then all high-to-low frequency-wise components could be transferred losslessly throughout the whole encoder-decoder process by symmetric multi-frequency parallel Transformer. Besides, the interaction of all high-to-low frequency-wise components are gradually enhanced by cross-frequency connection strategy and SHCP with multi-frequency fusion decoder (MFD) to generate the initial result.

(2) Mask-and-geo-knowledge-based uncertainty diffusion module (MUDM) is proposed to optimize the edge accuracy of the above initial map with the self-supervised learning strategy. Radiometric noise samples are generated based on multiple geo-knowledge including edge mixed pixels, edge occlusion, etc. And the certain regions of the initial map are masked by defining the uncertain regions according to the segmentation probability and edge uncertainty. Then, the certain regions are further optimized by the proposed MUDM to recover the labels with aforementioned noises.

(3) Semi-pseudo-Siamese UDHF$^2$-Net to perform change detection task is proposed with the self-supervised learning strategy. It is proposed to extract high-fidelity frequency-wise spatially-stationary-and-non-stationary features for generating the initial change map at a low computational cost. Besides the above geo-knowledge radiometric noises, the geometric registration error could be generated. Then uncertain regions of the initial change map are optimized by the proposed MUDM to obtain the final result.

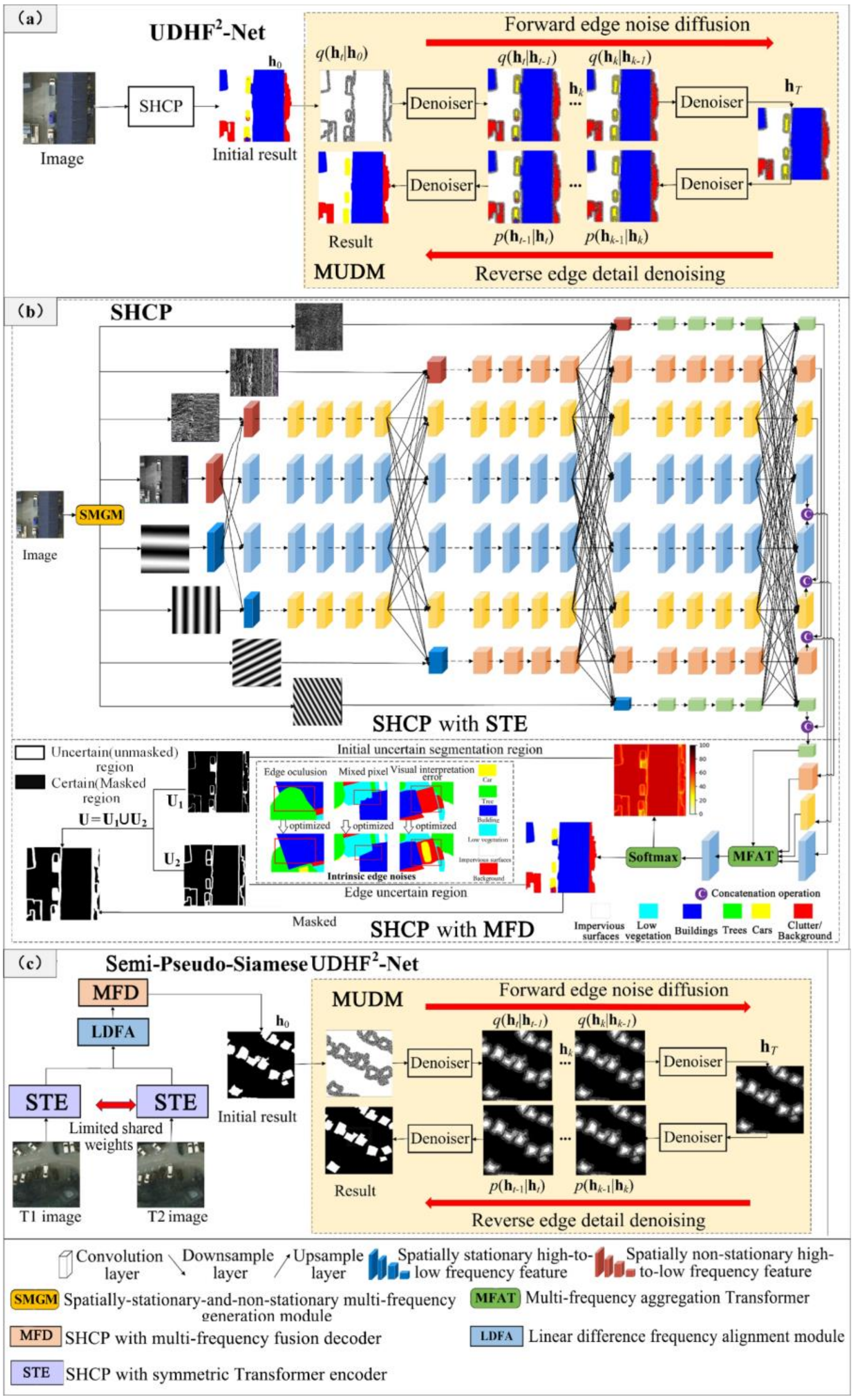

**Figure 2**. Overview of the proposed UDHF$^2$-Net. (a) the overall architecture; (b) spatially-stationary-and-non-stationary high-frequency connection paradigm (SHCP); and (c) the semi-pseudo-Siamese UDHF$^2$-Net for change detection task.

### *3.1 Spatially-stationary-and-non-stationary high-frequency connection paradigm (SHCP)*

As illustrated in Fig.2(b), SHCP adopts the encoder-decoder architecture to generate initial segmentation result, which consists of the following parts: (1) SHCP with symmetric Transformer encoder (STE); (2) SHCP with multi-frequency fusion decoder (MFD).

*3.1.1 SHCP with symmetric Transformer encoder (STE)*

The proposed encoder has four stages. In the first step, the input image $\mathbf{X} \in \mathbb{R}^{H \times W \times 3}$ could be transformed to multiple spatially frequency-wise stationary and non-stationary features $\mathbf{F}_{\text{stationary}}$ and $\mathbf{F}_{\text{non-stationary}}$ respectively, then they are fed into the SHCP with STE to generate the multiple high-to-low frequency-wise streams.

**(1) Spatially-stationary-and-non-stationary multi-frequency generation module (SMGM)**

SMGM is proposed to extraction spatially frequency-wise stationary and non-stationary features.

To capture local abrupt change information, $\mathbf{X}$ could be decomposed into four spatially frequency-wise non-stationary components based on Discrete Wavelet Transform along the rows and columns with the high-pass filter $\boldsymbol{F}_{\text{high}} = [1/\sqrt{2}, -1/\sqrt{2}]$ and the low-pass filter $\boldsymbol{F}_{\text{low}} = [1/\sqrt{2}, 1/\sqrt{2}]$, i.e., $\mathbf{F}_{\text{non-stationary}} = [\mathbf{T}_{\text{HH}}, \mathbf{T}_{\text{LH}}, \mathbf{T}_{\text{HL}}, \mathbf{T}_{\text{LL}}]$. The subband frequency-wise component $\mathbf{F}^{i}_{\text{non-stationary}}$ could been arranged from high-frequency to low-frequency as the index $i \in \{1,2,3,4\}$.

To capture global consistent information, the input $\mathbf{X}$ could be decomposed into the multiple high-to-low spatially frequency-wise stationary components with Discrete Fourier Transform, which is shown as Eq.(1):

$$\mathbf{F}_{\text{stationary}}(x,y) = \frac{1}{WH}\sum_{m=0}^{W-1}\sum_{n=0}^{H-1}\mathbf{X}(m,n)e^{-2l\pi(um/W+vn/H)}$$
$$= \frac{1}{WH}\sum_{m=0}^{W-1}\sum_{n=0}^{H-1}\mathbf{X}(m,n)(\cos 2\pi um/W - l\sin 2\pi um/W)(\cos 2\pi vn/H - l\sin 2\pi vn/H) \tag{1}$$

where $H$ and $W$ represent the height and width of the input image. $u$ and $v$ are the frequency coordinates. $\mathbf{F}_{\text{stationary}}$ represents the extracted multiple spatially frequency-wise stationary features which includes high-to-low frequency-wise features as the index $i \in \{1,2,3,4\}$.

Besides, multiple corresponding stride-2 3×3 convolution layers are performed for every frequency-wise component of $\mathbf{F}_{\text{stationary}}$ and $\mathbf{F}_{\text{non-stationary}}$ to decrease the resolution to 1/4, 1/8,1/16,1/32 of the original resolution respectively from high frequency to low frequency.

**(2) Symmetric multi-frequency parallel Transformer**

To enhance the discriminative independently of different frequency-wise features, the symmetric multi-frequency parallel transformer is proposed to improve the above spatially frequency-wise stationary and non-stationary feature representation respectively in every stage. Multi-scale frequency-wise components of $\mathbf{F}_{\text{stationary}}$ and $\mathbf{F}_{\text{non-stationary}}$ could be as the input to generate multiple parallel frequency-wise streams to provide fine-grained edge details and coarse-grained structure information. Starting from the first high-frequency stream, each frequency-wise component of $\mathbf{F}_{\text{stationary}}$ and $\mathbf{F}_{\text{non-stationary}}$ could be as input and gradually added to the next high-to-low frequency-wise streams over spatially frequency-wise stationary and non-stationary features in the next stage one by one. It is worth mentioning that the later symmetric frequency-wise streams always consist of all previous higher frequency-wise streams.

**(3) High-frequency Transformer module (HFTM)**

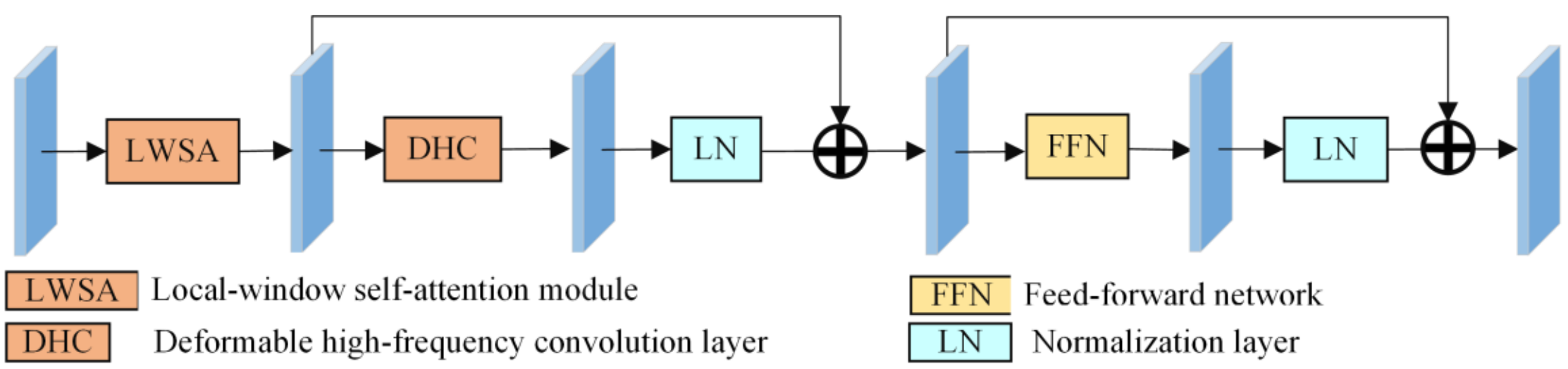

**Figure 3.** Illustration of the high-frequency Transformer module.

In every frequency-wise stream, the proposed high-frequency Transformer module (HFTM) illustrated in Fig.3 is proposed to introduce the deformable convolution for improving the adaptive frequency-wise aggregation ability with lager receptive field. For the input frequency-wise feature $\mathbf{Z} \in \mathbb{R}^{h \times w \times D}$ of each frequency-wise stream, this proposed module is composed of the following components: (1) Local-window self-attention module. Following formulations in HRFormer (Yuan et al., 2021), $\mathbf{Z}$ could be divided into a series of non-overlapping local windows of size $L\times L$: $\mathbf{Z} \Rightarrow \{\mathbf{Z}_1, \mathbf{Z}_2, \cdots, \mathbf{Z}_n\}$. Based on local-window self-attention module, the local frequency-wise features of each window are aggregated and merged to generate $\mathbf{M} \in \mathbb{R}^{h\times w\times E}$; (2) Feed-forward network with deformable high-frequency convolution layer is proposed and illustrated by:

$$\mathbf{D} = (\mathbf{M} + \mathrm{LN}(\mathrm{DHC}(\mathbf{M})) + \mathrm{LN}(\mathrm{FFN}(\mathbf{M} + \mathrm{LN}(\mathrm{DHC}(\mathbf{M})))\tag{2}$$

where $\mathrm{DHC}(\cdot)$ denotes deformable high-frequency convolution layer, $\mathrm{LN}(\cdot)$ denotes the normalization layer, $\mathrm{FFN}(\cdot)$ denotes Feed-forward network. It is worth mentioning that deformable high-frequency convolution layer could be formulated by (Wang et al., 2023a):

$$\mathbf{D}(r_0) = \sum_{t=1}^{T}\sum_{n=1}^{N}\big(\mathbf{W}_t\mathbf{U}_{tn}\mathbf{M}_t(r_0 + r_n + \triangle r_{tn})\big)\tag{3}$$

where the input $\mathbf{M}$ could be sliced into $T$ groups of feature map $\mathbf{M}_t \in \mathbb{R}^{h\times w\times E'}$, and $E' = E/T$ denotes the channel value of the group. $\mathbf{W}_g \in \mathbb{R}^{E'\times E}$ and $\mathbf{U}_{tn} \in \mathbb{R}$ represent the location-irrelevant weight matrix in the $t$-th group respectively. $\mathbf{U}_{tn} \in \mathbb{R}$ could be regarded as the modulation scalar of the $n$-th sampling location, which could be normalized by the softmax along the channel $N$. $\triangle r_{tn}$ denotes the offset of the sampling location $r_n$ of the $t$-th group.

**(4) Cross-frequency connection strategy**

To enhance the interaction and complementarity of multiple spatially frequency-wise stationary and non-stationary features, the cross-frequency connection strategy is proposed among different stages.

For instance, Fig.4 describes how to connect the multiple input spatially frequency-wise non-stationary features $\{\mathbf{N}_k^{\mathrm{I}}, k = 1{,}2\}$ and spatially frequency-wise stationary features $\{\mathbf{S}_k^{\mathrm{I}}, k = 1{,}2\}$ in details, where $k$ denotes the frequency index. And the related output features are $\{\mathbf{S}_k^{\mathrm{O}}, k = 1{,}2{,}3\}$. It is worth mentioning that every output frequency feature could fuse the high-to-low frequency-wise spatially-stationary-and-non-stationary features from the previous stage by: $\mathbf{S}_k^{\mathrm{O}}=\mathcal{H}_{1k}(\mathbf{N}_1^{\mathrm{I}})+\mathcal{H}_{2k}(\mathbf{N}_2^{\mathrm{I}})+\mathcal{H}_{3k}(\mathbf{N}_3^{\mathrm{I}})+\mathcal{H}_{1k}(\mathbf{S}_1^{\mathrm{I}})+\mathcal{H}_{2k}(\mathbf{S}_2^{\mathrm{I}})+\mathcal{H}_{3k}(\mathbf{S}_3^{\mathrm{I}})$.

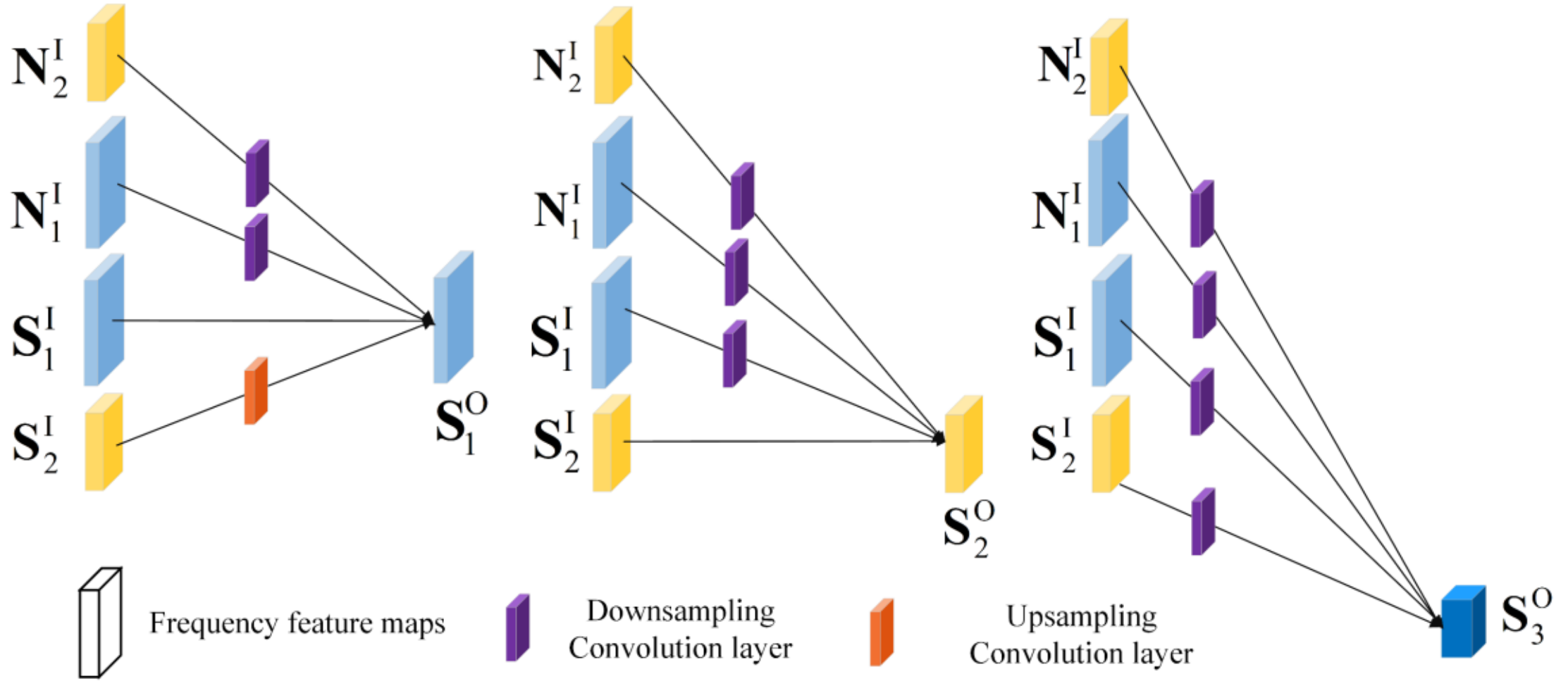

**Figure 4**. Illustration of the cross-frequency connection strategy. $\mathbf{N}_k^{\mathrm{I}}$ and $\mathbf{S}_k^{\mathrm{I}}$ represent input spatially frequency-wise non-stationary and stationary features. $\mathbf{S}_k^{\mathrm{O}}$ is the output of the spatially frequency-wise stationary feature. And $k$ represents the frequency index.

The connection function $\mathcal{H}_{jk}(\cdot)$ is determined by the input frequency index $j$. If $j=k$, $\mathcal{H}_{jk}(\mathbf{S})=\mathbf{S}$; If $j<k$, $\mathcal{H}_{jk}(\mathbf{S})=\mathrm{DWConv}(\mathbf{S})$ where $\mathrm{DWConv}(\cdot)$ denotes the downsampling operation by the $(k-j)$ stride-2 3×3 convolution layer; If $j>k$, $\mathcal{H}_{jk}(\mathbf{S})=\mathrm{UPConv}(\mathbf{S})$ where $\mathrm{UPConv}(\cdot)$ denotes the 1×1 bilinear upsampling convolution layer and the scale factor is set to $(k-j)$;

*3.1.2 SHCP with multi-frequency fusion decoder (MFD)*

As illustrated in Fig.2 (b), SHCP with MFD is proposed to adaptively align and fuse the above extracted spatially frequency-wise stationary and non-stationary features, and a multi-frequency aggregation Transformer (MFAT) is proposed to gradually enhance the fused frequency-wise features from low frequency to high frequency. The details are described as follows:

Firstly, the obtained spatially stationary and non-stationary multi-frequency features from the proposed encoder, i.e., $\mathbf{E}_{\text{stationary}}$ and $\mathbf{E}_{\text{non-stationary}}$, are fused by:

$$\mathbf{E}_{\text{fusion}}^{i}=\mathrm{Cat}\left(\mathbf{E}_{\text{non-stationary}}^{i},\mathbf{E}_{\text{stationary}}^{i}\right), i\in\{1,2,3,4\} \tag{4}$$

where $\mathbf{E}_{\text{fusion}}^{i}$ and $\mathrm{Cat}(\cdot)$ denote the $i$-th output fused frequency-wise feature and the concatenation operation respectively.

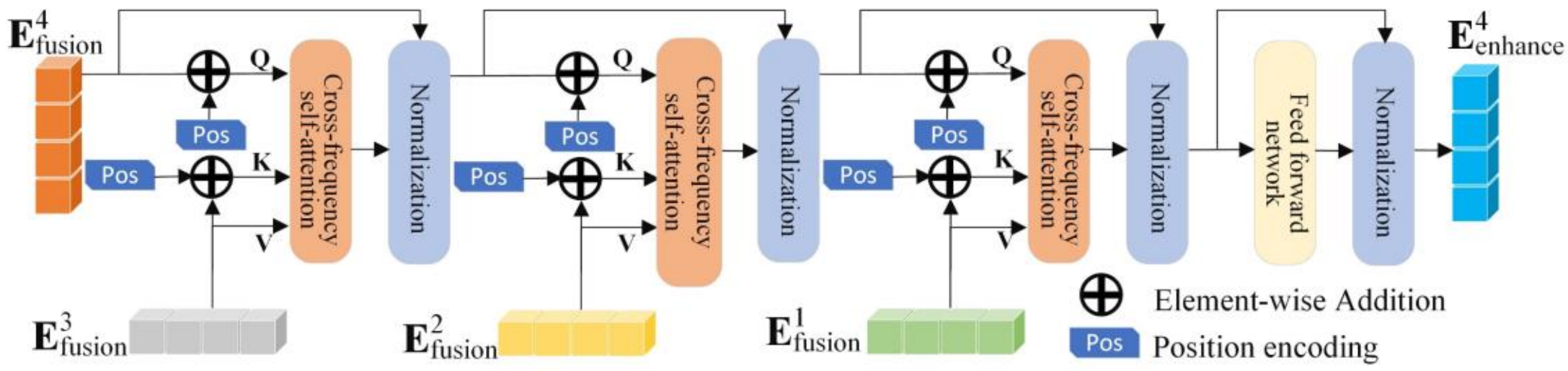

**Figure 5.** Illustration of the multi-frequency aggregation Transformer (MFAT).

Secondly, MFAT is leveraged to model high-frequency and low-frequency correlation for enhancing feature representation as shown in Fig.5, aiming at taking the complementary advantage of the edge information in high-frequency features and structural information in low-frequency features. For example, $\mathbf{E}^4_{\text{fusion}}$ could be strengthened from low frequency to high frequency by gradually fusing $\mathbf{E}^3_{\text{fusion}}, \mathbf{E}^2_{\text{fusion}}, \mathbf{E}^1_{\text{fusion}}$, which is described by:

$$\mathbf{E}^4_{\text{enhance}} = \text{NL}\left(\text{FFN}\left(\text{NL}\left(\text{CS}\left(\text{NL}\left(\text{CS}\left(\text{NL}\left(\text{CS}\left(\mathbf{T}^4_{\text{fusion}}, \mathbf{T}^3_{\text{fusion}}\right)\right), \mathbf{T}^2_{\text{fusion}}\right)\right), \mathbf{T}^1_{\text{fusion}}\right)\right)\right)\right) \quad (5)$$

where $\mathbf{E}^1_{\text{enhance}}$, $\text{CS}(\cdot)$, $\text{NL}(\cdot)$ and $\text{FFN}(\cdot)$ represent the enhanced output feature, the cross-frequency self-attention layer, normalization layer and Feed-forward network.

The cross-frequency self-attention layer could be performed by efficient attention (Zhuoran et al., 2021) to improve multi-frequency interaction. Different frequency-wise features could be projected into tokens with the same size, the input low-frequency features are linearly transformed into $\mathbf{Q}$ value, the input high-frequency features are linearly transformed into $\mathbf{V}$ and $\mathbf{K}$ values. The cross-frequency self-attention layer is leveraged to compute the attention weights for capturing the correlation of different frequency-wise features.

Finally, the fusion features could be gradually enhanced to generate $\mathbf{E}^1_{\text{enhance}}$. And a 4×upsampling convolution layer is added to recover deep features to the original resolution.

*3.2 Mask-and-geo-knowledge-based uncertainty diffusion module (MUDM)*

Although the above proposed SHCP could effectively alleviate the edge detail loss,

the inherent uncertainties in the edge still exist including the following radiometric noises: (1) mixed pixels; (2) edge occlusion; (3) visual interpretation errors. To further address these noises, MUDM is proposed to adopt a self-supervised learning strategy to optimize the edge of the initial segmentation result, which could automaticly generate the samples with denoising diffusion probabilistic model, including the images and labels with edge noises and registration error, for semantic segmentation and change detection task. And denoising diffusion probabilistic model is leveraged to overcome the impact of edge noises and recover to the truth labels. Therefore, MUDM is proposed as follows:

(1) Masking the certain region by finding the uncertain region. The initial segmentation result of SHCP could be defined as two regions, i.e., the certain and uncertain regions. The uncertain region consists of the following two parts in Fig.2 (b):

① Defining the initial uncertain (unmasked) segmentation region. It is defined by the segmentation probability of SHCP. If the maximum segmentation probabilities of these pixels are lower than a threshold $\rho$, these pixels could be set as the uncertain region. The formula is defined by:

$$\mathbf{U} = \begin{cases} 1, \text{if } \max\left(e^{O_k^1}/\sum_{i=1}^{N} e^{O_k^i}, \ldots, e^{O_k^j}/\sum_{i=1}^{N} e^{O_k^i}, \ldots, e^{O_k^N}/\sum_{i=1}^{N} e^{O_k^i}\right) \leq \rho \\ 0, \text{ otherwise} \end{cases} \quad (6)$$

Where $\mathbf{U}$ is defined as the initial uncertain segmentation region based on the segmentation probability, $O_k^j$ is the output of the $k$-th pixel from the proposed SHCP for the $j$-th category, $N$ represents the category number, $\max(\cdot)$ is used to obtain the maximum probability. Notably, $\mathbf{U}$ includes two types of uncertain regions: edge uncertain regions and the other uncertain regions (i.e., mixed pixels and ambiguous ground objects).

② Masking the certain region. The certain region $\mathbf{C}$ are obtained by:

$$\mathbf{C} = \mathbf{R} - \mathbf{U} \quad (7)$$

Where $\mathbf{R}$ is the initial segmentation label, $\mathbf{C}$ is defined as the certain region. $\mathbf{C}$ could be masked to remain the initial segmentation and $\mathbf{U}$ could be optimized

Importantly, the certain region $\mathbf{C}$ is masked to preserve the initial segmentation result, while the uncertain (unmasked) region $\mathbf{U}$ could be further optimized by the proposed MUDM.

(2) Generating negative sample image with geo-knowledge-based radiometric noises. Two strategies of data augmentation for input images and corresponding labels are proposed to improve the anti-noise ability as follows: ① edge mixed pixel augmentation. In the uncertain region, the mixed pixel $M^n$ for the $n$-th band of image (e.g., $n \in [1,4]$) could be constructed from three samples $M_i^n$, $M_j^n$ and $M_k^n$ of different classes:

$$M^n = \alpha M_i^n + \beta M_j^n + (1 - \alpha - \beta) M_k^n, s.t. 0 \leq \alpha, \beta \leq 1 \tag{8}$$

where $\alpha$, $\beta$ are the weights. $M_i^n$, $M_j^n$ and $M_k^n$ respectively represent the random pixels of the $i$-th, $j$-th and $k$-th class which respectively represent impervious surfaces, buildings and clutter/background in ISPRS Vaihingen and Potsdam 2D datasets; ② edge occlusion sample augmentation. The incorrect occlusion labels are corrected and massive generated occlusion samples are generated. Occlusion samples are chosen and randomly located in the range of the uncertainty edge regions to overlay on the occluded samples to increase the variability of samples.

(3) Optimizing the uncertain segmentation with MUDM by a self-supervised learning strategy. MUDM is proposed to optimize the above uncertain regions in Fig.2 (a), which consists of the following two processes:

In the forward edge noise diffusion process, MUDM is self-supervised learning strategy to train a Markovian noising process, which could simulate data distribution of the above given a series of noisy knowledge from the uncertain region of the initial segmentation results (i.e., labels) $\mathbf{h}_0$ to generate noise samples (i.e., labels) $\mathbf{h}_T$. It gradually adds the geo-knowledge-based noises from data distribution $q(\mathbf{h}_0)$ to $\mathbf{h}_T$ over a finite iteration $T$ based on a variance schedule $\mu_1, \ldots\ldots, \mu_T$, which could be formulated by:

$$\begin{gathered} q(\mathbf{h}_{1:T}|\mathbf{h}_0) = \prod_{t=1}^{T} q(\mathbf{h}_t|\mathbf{h}_{t-1}) \\ q(\mathbf{h}_t|\mathbf{h}_{t-1}) = \mathcal{N}\left(\mathbf{h}_t; \sqrt{1-\mu_t}\mathbf{h}_{t-1}, \mu_t \mathbf{I}\right) \end{gathered} \tag{9}$$

where $\mathcal{N}(\cdot;\cdot,\cdot)$ satisfies the Gaussian distribution, $\mu_t\mathbf{I}$ and $\mathbf{I}$ represent the variance matrix and the identity matrix. standard normal distribution. And the direct data distribution $q(\mathbf{h}_t|\mathbf{h}_0)$ from $\mathbf{h}_0$ to $\mathbf{h}_t$ could be derived by:

$$q(\mathbf{h}_t|\mathbf{h}_0) = \mathcal{N}\left(\mathbf{h}_t; \sqrt{\gamma_t}\mathbf{h}_0, (1-\gamma_t)\mathbf{I}\right) = \sqrt{\gamma_t}\mathbf{h}_0 + \sqrt{1-\gamma_t}\boldsymbol{\varphi}, \boldsymbol{\varphi} \sim \mathcal{N}(0, \mathbf{I}) \quad (10)$$

where $\gamma_t = 1 - \mu_t$ and $\gamma_t = \prod_{k=1}^{t} \gamma_k$ , and $\boldsymbol{\varphi}$ indicates random noise from a Gaussian distribution in semantic segmentation.

In the reverse edge detail denoising process, MUDM could leverage the denoiser to gradually remove the noises of $\mathbf{h}_T$ and recover to the correct label from the uncertain result of initial segmentation. The proposed SHCP is chosen as denoiser to parameterize the reverse distribution. It is performed by:

$$p_\theta(\mathbf{h}_{t-1}|\mathbf{h}_t) := \mathcal{N}(\mathbf{h}_{t-1}: \nu_\theta(\mathbf{h}_t, t), \textstyle\sum_\theta(\mathbf{h}_t, t)) \quad (11)$$

where $t \in [1, T]$, $\nu_\theta$ and $\sum_\theta$ represent the mean and variance of the prior distribution segmented by the model.

*3.3 Semi-Pseudo-Siamese UDHF$^2$-Net for change detection*

Semi-pseudo-Siamese UDHF$^2$-Net is proposed to further reduce false change detection caused by image registration error of bi-temporal images for change detection task in Fig.2 (c). It is a variant of UDHF$^2$-Net to specifically address change detection including all the modules of the original UDHF$^2$-Net, which consists of the following two stages:

(1) Obtaining initial changed label

In the decoder phase, the semi-pseudo-Siamese architecture with partially shared weights is adopted as the encoder for the input of $\mathbf{X}_1 \in \mathbb{R}^{H\times W\times 3}$ and $\mathbf{X}_2 \in \mathbb{R}^{H\times W\times 3}$. It is beneficial to independently extract frequency-wise spatially-stationary-and-non-stationary high-to-low features from bi-temporal images by:

$$(\mathbf{P}_{\text{stationary}}, \mathbf{P}_{\text{non-stationary}}), \left(\mathbf{Q}_{\text{stationary}}, \mathbf{Q}_{\text{non-stationary}}\right) = \text{STE}(\mathbf{X}_1), \text{STE}(\mathbf{X}_2) \quad (12)$$

where $\mathbf{P}_{\text{stationary}}$ and $\mathbf{Q}_{\text{stationary}}$ denote the output multiple spatially frequency-wise stationary features of $\mathbf{X}_1$ and $\mathbf{X}_2$ , $\mathbf{P}_{\text{non-stationary}}$ and $\mathbf{Q}_{\text{non-stationary}}$ denote the output multiple spatially frequency-wise non-stationary features of $\mathbf{X}_1$ and $\mathbf{X}_2$, $\text{STE}(\cdot)$ represents the proposed SHCP with STE in section 3.1.1. Different from the pseudo-Siamese architecture, the semi-pseudo-Siamese architecture has shared most parameters and remains the last two layers to independently train, which could effectively reduce computational memory.

To adaptively align difference features of multiple frequency-wise features, a linear difference frequency alignment module (LDFA) is proposed by:

$$\begin{aligned}\mathbf{R}^{i}_{\mathrm{stationary}} &= \mathrm{BN}\left(\mathrm{Relu}\left(\mathrm{Conv}\left(\mathrm{Cat}\left(\mathbf{P}^{i}_{\mathrm{stationary}}, \mathbf{Q}^{i}_{\mathrm{stationary}}\right)\right)\right)\right)\\ \mathbf{R}^{i}_{\mathrm{non-stationary}} &= \mathrm{BN}\left(\mathrm{Relu}\left(\mathrm{Conv}\left(\mathrm{Cat}\left(\mathbf{P}^{i}_{\mathrm{non-stationary}}, \mathbf{Q}^{i}_{\mathrm{non-stationary}}\right)\right)\right)\right)\end{aligned} \tag{13}$$

where $\mathrm{Cat}(\cdot)$, $\mathrm{Relu}(\cdot)$, $\mathrm{Conv}(\cdot)$, $\mathrm{Relu}(\cdot)$, $\mathrm{BN}(\cdot)$, $\mathbf{R}^{i}_{\mathrm{non-stationary}}$, $\mathbf{R}^{i}_{\mathrm{stationary}}$ denote the concat operation, Relu activation function, 1×1 convolution layer, batch normalization layer, the *i*-th output of spatially frequency-wise non-stationary features and the *i*-th output of spatially frequency-wise stationary features respectively.

In decoder phase, the proposed SHCP with MFD is chosen as the decoder to enhance the complementarity by:

$$\mathbf{O} = \mathrm{MFD}\left(\mathbf{F}_{\mathrm{non-stationary}}, \mathbf{F}_{\mathrm{stationary}}\right) \tag{14}$$

where $\mathbf{O}$ is the output of the initial change label and $\mathrm{MFD}(\cdot)$ represents the SHCP with MFD in subsection 3.1.2.

(2) Optimizing uncertain region

The proposed MUDM in section 3.2 is also adopted to optimize the initial change result. It consists of the following parts:

① Masking the certain region by finding the uncertain region. Based on the above obtained initial change label, the certain region is masked to remain the original result, and the rest is regarded as the uncertain (unmasked) region. The detailed description is shown in section 3.2.

② Generating negative sample image with geo-knowledge-based noises. Apart from simulating radiometric noises as section 3.2, the false change images are generated by registration error:

$$
\begin{aligned}
\acute{x} &= a_0 + \Delta a_0 + (a_1 + \Delta a_1)x + (a_2 + \Delta a_2)y \\
\acute{y} &= b_0 + \Delta b_0 + (b_1 + \Delta b_1)x + (b_2 + \Delta b_2)y
\end{aligned} \tag{15}
$$

where $x$ and y represent the coordinate of the image on time phase 2, $\acute{x}$ and $\acute{y}$ represent the generated coordinate of the image on time phase 2 with random registration error, $a_0, a_1, a_2, b_0, b_1, b_2$ are the random correction coefficients, where $\Delta a_0, \Delta b_0 \in [-1.5,1.5]$; and $|\Delta a_1|, |\Delta a_2|, |\Delta b_1|, |\Delta b_2| \in [10^{-3}, 10^{-5}]$.

③ Optimizing the uncertain change with uncertainty diffusion model. The uncertain region of the initial change segmentation result is optimized by uncertainty diffusion model as the section 3.2 to improve the edge extraction accuracy.

### *3.4 Loss function*

#### *3.4.1 Loss function of semantic segmentation*

The two-stage cascade loss is leveraged to optimize the proposed UDHF²-Net in semantic segmentation task.

In the first stage, a hybrid loss $\mathcal{L}_{\text{seg}}$ is applied to train the proposed SHCP for achieving the initial segmentation result, which consists of a dice loss $\mathcal{L}_{\text{dice}}$ and a cross-entropy loss $\mathcal{L}_{\text{ce}}$. It could be formulated as by:

$$
\begin{aligned}
\mathcal{L}_{\text{seg}} &= \gamma \mathcal{L}_{\text{ce}} + (1-\gamma)\mathcal{L}_{\text{dice}} \\
\mathcal{L}_{\text{ce}} &= -\left(\textstyle\sum_{n=1}^{N}\sum_{c=1}^{C}(y^{c,n}\log\hat{y}^{c,n})\right)/N \\
\mathcal{L}_{\text{dice}} &= 1 - 2\left(\textstyle\sum_{n=1}^{N}\sum_{c=1}^{C}(y^{c,n}\hat{y}^{c,n}/(y^{c,n}+\hat{y}^{c,n}))\right)/N
\end{aligned} \tag{16}
$$

where $\gamma \in [0,1]$ is the weighted coefficient, $C$ is the number of classes and $N$ is the total number of pixels in the image. $y$ represents the one-hot encoding of the truth label. $\hat{y}$ is the output predicted probability of the SHCP.

In the second stage, the uncertain loss $\mathcal{L}_{\text{uncertain}}$ is leveraged to optimize the uncertain region of the initial segmentation result, which consists of an edge uncertainty loss $\mathcal{L}_{E\text{u}}$ and a diffusion loss $\mathcal{L}_{\text{diff}}$ by:

$$
\begin{aligned}
\mathcal{L}_{\text{uncertain}} &= \omega \mathcal{L}_{\text{diff}} + (1-\omega)\mathcal{L}_{E\text{u}} \\
\mathcal{L}_{\text{diff}} &= \mathbb{E}_{t,\mathbf{h}_0,\varphi}\left[\| \boldsymbol{\varphi}_{\text{seg}} - \boldsymbol{\varphi}_{\text{seg}}^{\theta}(\sqrt{\bar{\mu}_t}\mathbf{h}_0 + \sqrt{1-\bar{\mu}_t}\boldsymbol{\varphi}_t) \|^2\right] \\
\mathcal{L}_{E\text{u}} &= 1 - 2 \times \text{Precison}_{\text{seg}} \times \text{Recall}_{\text{seg}} / (\text{Precison}_{\text{seg}} + \text{Recall}_{\text{seg}})
\end{aligned} \tag{17}
$$

where $\omega \in [0,1]$ is the weighted coefficient, $\boldsymbol{\varphi}_{\text{seg}}^{\theta}$ indicates the parameterized denoiser, $\boldsymbol{\varphi}_{\text{seg}} \sim \mathcal{N}(0, \mathbf{I})$ indicates random noise from a Gaussian distribution in semantic segmentation, and $\mathbf{h}_0$ represents the initial segmentation result. $\boldsymbol{\varphi}_t$ is a function approximator aimed to predict $\boldsymbol{\varphi}$ from $\mathbf{h}_t$ $\mathrm{Precison}_{\mathrm{seg}}$ and $\mathrm{Recall}_{\mathrm{seg}}$ are the precision and recall values to evaluate the truth label and predicted label in the uncertain region.

*3.4.2 Loss function of change detection*

The two-stage cascade loss is used to optimize the proposed semi-pseudo-Siamese UDHF$^2$-Net in change detection task.

In the first stage, the hybrid loss $\mathcal{L}_{\mathrm{cd}}$ is chosen to optimize the initial change result, which consists of a weighted cross entropy loss $\ell_{\mathrm{wbce}}$ and a dice loss $\ell_{\mathrm{dice}}$ by:

$$\begin{gathered}\ell_{\mathrm{cd}} = g\ell_{\mathrm{wbce}} + (1-g)\ell_{\mathrm{dice}} \\ \ell_{\mathrm{wbce}} = -(\lambda\varrho\log(\hat{\varrho}) + (1-\lambda)(1-\varrho)\log(1-\hat{\varrho})) \\ \ell_{\mathrm{dice}} = 1 - 2\varrho\hat{\varrho}/(\varrho+\hat{\varrho})\end{gathered} \tag{18}$$

where $g, \lambda \in [0,1]$ is the weighted coefficient, $\varrho_i$ represents the truth label and $\hat{\varrho}$ is the predicted probability.

In the second stage, the uncertain loss $\ell_{\mathrm{uncertain}}$ is chosen to optimize the uncertain region of the initial change result, which consists of an edge uncertainty loss $\ell_{\mathrm{Eu}}$ and a diffusion loss $\ell_{\mathrm{diff}}$ by:

$$\begin{gathered}\ell_{\mathrm{uncertain}} = \lambda\ell_{\mathrm{diff}} + (1-\lambda)\ell_{\mathrm{Eu}} \\ \ell_{\mathrm{diff}} := \mathbb{E}_{t,\mathbf{Y}_0,\psi}\left[\| \boldsymbol{\varphi}_{\mathrm{cd}} - \boldsymbol{\varphi}_{\mathrm{cd}}^{\theta}(\sqrt{\bar{\mu}_t}\mathbf{Y}_0 + \sqrt{1-\bar{\mu}_t}\boldsymbol{\psi}_t) \|^2\right] \\ \ell_{\mathrm{Eu}} = 1 - 2 \times \mathrm{Precison}_{\mathrm{cd}} \times \mathrm{Recall}_{\mathrm{cd}}/(\mathrm{Precison}_{\mathrm{cd}} + \mathrm{Recall}_{\mathrm{cd}})\end{gathered} \tag{19}$$

where $\lambda \in [0,1]$ is the weighted coefficient, $\boldsymbol{\varphi}_{\mathrm{cd}}^{\theta}$ indicates the parameterized denoiser, $\boldsymbol{\varphi}_{\mathrm{cd}} \sim \mathcal{N}(0, \mathbf{I})$ indicates random noise from a Gaussian distribution in change detection, and $\mathbf{Y}_0$ represents the initial change detection result. $\boldsymbol{\psi}_t$ is a function approximator aimed to predict $\boldsymbol{\psi}$ from $\mathbf{Y}_t$. $\mathrm{Precison}_{\mathrm{cd}}$ and $\mathrm{Recall}_{\mathrm{cd}}$ are the precision and recall values to evaluate the truth label and predicted label in the uncertain region.

***3.5 Evaluation metrics***

The following metrics are chosen: overall accuracy (OA), F1 score (F1), mean F1 score (mF1) and mean intersection over union (mIoU) for semantic segmentation task; Precision, Recall, F1 and intersection over union (IoU) for change detection task.

## 4. Experiments and discussion

To verify the superiorities of UDHF$^2$-Net in semantic segmentation and change detection tasks of RSII, the following experiments were conducted to compare with the state-of-the-art (SOTA) networks on several public benchmark datasets.

*4.1 Experimental setting*

(1) Experimental datasets

Excellent and remarkable public benchmark datasets for remote sensing semantic segmentation and change detection including the available largest and newest datasets are leveraged to evaluate the proposed networks. The datasets are described as follows:

① Semantic segmentation datasets

**ISPRS Vaihingen 2D Dataset (Vaihingen dataset).** This dataset was collected by the International Society for Photogrammetry and Remote Sensing from Vaihingen, which has four bands (i.e., near-infrared, red, and green bands). It consists of 33 true orthophoto images, one of which has the size of 2494×2064 pixels with ground sampling distance of 9 cm. It is divided into training, validation and testing datasets, including 17, 1 and 15 images respectively. This dataset is cut into the same size patches of 512×512 pixels with a sliding window striding 256 pixels. It consists of six categories, namely, impervious surfaces, buildings, low vegetation, trees, cars, and clutter/background. It is worth noting that some occluded buildings by low vegetation or trees could be corrected and recovered.

**ISPRS Potsdam 2D Dataset (Potsdam dataset).** This dataset was collected by the International Society for Photogrammetry and Remote Sensing from a 1.38 km$^2$ area of Potsdam with four bands (i.e., near-infrared, red, green and blue bands). It consists of 38 true orthophoto images, one of which has the size of 6000×6000 pixels with ground sampling distance of 9 cm. This dataset is divided into training, validation and testing datasets, including 23, 1 and 14 images respectively. It is cut into the same size patches of 512×512 pixels using a sliding window striding 256 pixels. Six categories are included in this dataset as the same as the ISPRS Vaihingen 2D Dataset. Notably some occluded buildings by low vegetation or trees could be corrected and recovered.

**LoveDA Dataset.** This dataset was collected from Nanjing, Changzhou, and Wuhan in China covering urban and rural scene. It contains 5987 images with a high-resolution 0.3 m, one of which has the size of 1024×1024 pixels. The images are divided into training, validation, and testing sets, including 2522, 1669, and 1796 images respectively. These images are annotated with seven land-cover categories: building, road, water, barren, forest, agriculture, and background.

② Change detection datasets

**WHU building change detection dataset (WB-CD dataset).** This dataset was sampled from two cities, i.e., Christchurch and New Zealand in 2012 and 2016. It consists of a pair image with a size of 15354 × 32507 pixels at a spatial resolution of 0.2 m. The image pairs were cut into non-overlapping pair samples with a size of 256 × 256 pixels to generate the training, validation and test datasets in the ratio of 1:1:8.

**LEVIR-CD dataset.** This dataset is a collection of building dataset obtained from Google Earth imagery with a high spatial resolution of 0.5 m. It includes 31333 pairs of change instances, which ware divided into 10192 pairs of image patches with a size of 256 × 256 pixels to generate 7120 pairs of non-overlapping images for training, 1024 pairs of images for validating and 2048 pairs of images for testing.

**SYSU-CD dataset.** This dataset was collected in Hong Kong between 2007 and 2014. It comprises a total of 20,000 pairs of non-overlapping image patches, which are extracted from high-resolution satellite images with a spatial resolution of 0.5 m. And each patch has the size of 256×256 pixel. The SYSU-CD dataset is divided into training, validation, and testing sets, with 12,000, 4,000, and 4,000 image pairs, respectively.

(2) Comparison networks

To confirm the effectiveness of the proposed UDHF$^2$-Net, several comparative DNNs for semantic segmentation task were chosen as the following: MAResU-Net (Li et al., 2022), SwiftNet (Wang et al., 2021a), ABCNet (Li et al., 2021), Segmenter (Strudel et al., 2021), BANet (Wang et al., 2021b), Swin Transformer (Liu et al., 2021), FT-UNetFormer (Wang et al., 2022a), DeepLabV3+ (Chen et al., 2017), PSPNet (Zhao et al., 2017), LANet (Ding et al., 2021), MANet (Niu et al., 2021), HRNetV2 (Wang et al., 2020), HRFormer (Yuan et al., 2021), EaNet (Zheng et al., 2020), SFFNet (Yang et al., 2024), SkySense (Guo et al., 2024), CMID (Muhtar et al., 2023), TOV (Tao et al., 2023), GFM (Mendieta et al., 2023), Ringmo-SAM (Yan et al., 2023), RSProtoSeg (Sun et al., 2024), CAGNet (Wang et al., 2024b) and SSG2 (Diakogiannis et al., 2024).

To evaluate the advancement of the proposed semi-pseudo-Siamese UDHF$^2$-Net, some excellence DNNs were compared for change detection task as following: AMTNet (Liu et al., 2023), FC-EF (Caye Daudt et al., 2018), FC-Siam-conc (Caye Daudt et al., 2018), FC-Sima-diff (Caye Daudt et al., 2018), DTCDSCN (Liu et al., 2020), SNUNet (Fang et al., 2022), BiT (Chen et al., 2022),, MTCNet (Wang et al., 2022b), GAS-Net (Zhang et al., 2023), W-Net (Wang et al., 2023b), ChangerEX (Fang et al., 2023), P2V-CD (Lin et al., 2023), ChangeFormer (Bandara and Patel, 2022), SGSLN (Zhao et al., 2023), IFN (Zhang et al., 2020), MFPNet (Xu et al., 2021), ChangeCLIP (Dong et al., 2024), MSP-CD (Ning et al., 2024), SkySense (Guo et al., 2024), GFM (Mendieta et al., 2023) and CMID (Muhtar et al., 2023).

**Table 1.** Quantitative analysis results on the ISPRS Potsdam 2D test dataset with compared SOTA DNNs. The highest values are highlighted in bold for every evaluation metric (%). (Some values are null because they were not provided in the original paper.)

| Method | Backbone | F1 | | | | | MeanF1 | OA | mIoU |
|---|---|---|---|---|---|---|---|---|---|
| | | Imp.surf. | Building | Low.veg. | Tree | Car | | | |
| SwiftNet | ResNet18 | 91.83 | 95.94 | 85.72 | 86.84 | 94.46 | 90.96 | 89.33 | 83.84 |
| ABCNet | ResNet18 | 93.50 | 96.90 | 87.90 | 89.10 | 95.80 | 92.70 | 91.30 | 86.50 |
| Segmenter | ViT-Tiny | 91.50 | 95.30 | 85.40 | 85.00 | 88.50 | 89.20 | 88.70 | 80.70 |
| BANet | ResT-Lite | 93.34 | 96.66 | 87.37 | 89.12 | 95.99 | 92.50 | 91.06 | 86.25 |
| Swin Transformer | Swin-Tiny | 93.20 | 96.40 | 87.60 | 88.60 | 95.40 | 92.20 | 90.90 | 85.80 |
| FT-UNetFormer | Swin-Base | 93.90 | 97.20 | 88.80 | 89.80 | 96.60 | 93.30 | 92.00 | 87.50 |
| DeepLabV3+ | ResNet18 | 90.53 | 95.89 | 83.61 | 84.25 | 88.69 | 88.59 | 87.97 | 80.56 |
| PSPNet | ResNet18 | 90.56 | 95.65 | 84.61 | 84.72 | 88.49 | 90.51 | 88.28 | 80.32 |
| LANet | ResNet18 | 93.05 | 97.19 | 87.30 | 88.04 | 94.19 | 91.95 | 90.84 | — |
| MANet | ResNet50 | 93.25 | 96.63 | 87.99 | 88.95 | 96.39 | 92.64 | 91.05 | 89.01 |
| SFFNet | — | 93.73 | 97.02 | 89.02 | 90.26 | 96.79 | 93.36 | 91.88 | 87.73 |
| HRNet | HRNetV2-W48 | 91.92 | 96.16 | 85.65 | 86.97 | 94.63 | 91.07 | 89.63 | 83.94 |
| TOV | — | — | — | — | — | — | 92.03 | — | — |
| CMID | ResNet50 | — | — | — | — | — | 92.81 | 92.74 | 87.04 |
| GFM | — | — | — | — | — | — | 91.85 | — | — |
| SkySense | — | — | — | — | — | — | 93.99 | — | — |
| Ringmo-SAM | — | — | — | — | — | — | — | — | 78.21 |
| RSProtoSeg | — | — | — | — | — | — | — | — | 75.5 |
| CAGNet | ResNet18 | 94.3 | 97.1 | 88.2 | 89.4 | 96.5 | 93.0 | 91.8 | 87.2 |
| HRFormer | — | 93.31 | 96.35 | 87.74 | 88.43 | 95.64 | 92.29 | 91.12 | 85.93 |
| SSG2 | — | 93.7 | 97.7 | 88.4 | 88.9 | 96.8 | 93.1 | 91.6 | 87.3 |
| Ours | — | **96.20** | **98.06** | **92.18** | **94.44** | **98.03** | **95.78** | **95.42** | **92.00** |

(3) Implementation details

All experiments were conducted using the PyTorch framework on four NVIDIA GTX 3090Ti GPU.

**Hyperparameter setting.** The AdamW optimizer was chosen to train all networks. Specially, we conducted the cosine strategy with an initial learning rate of 10-4 to adjust learning rate. The epoch was set to 100. The batchsize was set to 4 and 8 for semantic segmentation and change detection tasks respectively.

**Data augmentation strategy**. Data augmentation strategies were chosen such as random scaling ([0.5, 0.75, 1.0, 1.25, 1.5]), random vertical and horizontal flipping.

Besides, some strategies were conducted to effectively mitigate the risk of overfitting, such as early stopping, dropout, data augmentation and the selected largest and newest datasets.

*4.2 Semantic segmentation results*

*4.2.1 Comparison experiments of semantic segmentation*

**Table 2**. Quantitative analysis results on the ISPRS Vaihingen 2D test dataset for compared SOTA DNNs. The highest values are highlighted in bold for every evaluation metric (%). (Some values are null because they were not provided in the original paper.)

| Method | Backbone | F1 | | | | | MeanF1 | OA | mIoU |
|---|---|---|---|---|---|---|---|---|---|
| | | Imp.surf. | Building | Low.veg. | Tree | Car | | | |
| MAResU-Net | ResNet18 | 92.91 | 95.26 | 84.95 | 89.94 | 88.33 | 90.28 | 90.86 | 83.90 |
| SwiftNet | ResNet18 | 92.22 | 94.84 | 84.14 | 89.31 | 81.23 | 88.35 | 90.20 | 79.58 |
| ABCNet | ResNet18 | 92.70 | 95.20 | 84.50 | 89.70 | 85.30 | 89.50 | 90.70 | 81.30 |
| Segmenter | ViT-Tiny | 89.80 | 93.00 | 81.20 | 88.90 | 67.60 | 84.10 | 88.10 | 73.60 |
| BANet | ResT-Lite | 92.23 | 95.23 | 83.75 | 89.92 | 86.76 | 89.58 | 90.48 | 81.35 |
| Swin Transformer | Swin-Tiny | 92.80 | 95.60 | 85.10 | 90.60 | 85.10 | 89.80 | 91.00 | 81.80 |
| FT-UNetFormer | Swin-Base | 93.50 | 96.00 | 85.60 | 90.80 | 90.40 | 91.30 | 91.60 | 84.10 |
| EaNet | ResNet101 | 93.40 | 96.20 | 85.60 | 90.50 | 88.30 | 90.80 | 91.20 | — |
| DeepLabV3+ | ResNet18 | 91.35 | 94.34 | 81.32 | 87.42 | 78.14 | 86.60 | 88.91 | 75.34 |
| PSPNet | ResNet18 | 91.44 | 94.38 | 81.52 | 87.91 | 78.02 | 86.65 | 88.99 | 76.81 |
| LANet | ResNet18 | 92.41 | 94.90 | 82.89 | 88.92 | 81.31 | 88.09 | 89.83 | — |
| MANet | ResNet50 | 93.02 | 95.47 | 84.64 | 89.978 | 88.95 | 90.41 | 90.96 | 83.40 |
| SFFNet | — | 93.51 | 96.25 | 85.94 | 91.43 | 91.24 | 91.67 | 91.91 | 84.80 |
| HRNet | HRNetV2-W48 | 92.34 | 94.92 | 83.89 | 89.54 | 82.36 | 88.61 | 90.41 | 80.34 |
| HRFormer | HRFormer-B | 92.91 | 95.63 | 84.34 | 90.56 | 87.33 | 90.15 | 91.24 | 81.94 |
| GFM | — | — | — | — | — | — | — | — | 75.3 |
| Ringmo-SAM | — | — | — | — | — | — | — | — | 72.44 |
| CAGNet | — | 93.1 | 95.6 | 85.5 | 90.9 | 89.5 | 90.9 | 91.4 | 83.5 |
| RSProtoSeg | — | — | — | — | — | — | — | — | 70.7 |
| SSG2 | — | 92.9 | 95.2 | 84.3 | 89.8 | 85.6 | 89.6 | 90.7 | 81.4 |
| Ours | — | **95.13** | **97.62** | **86.98** | **92.24** | **93.58** | **93.11** | **93.57** | **87.31** |

As reported Table 1, 2 and 3, experimental results of the chosen benchmark datasets indicate that UDHF$^2$-Net achieves superior performance than the existing SOTA comparison networks. The proposed network yielded the highest mIOU values of 92.00%, 87.31% and 55.27% on Potsdam, Vaihingen, and LoveDA datasets respectively. It also achieved the superior OA values of 95.42% and 93.57%, and the highest MeanF1 95.78% and 93.11% on Potsdam and Vaihingen datasets respectively.

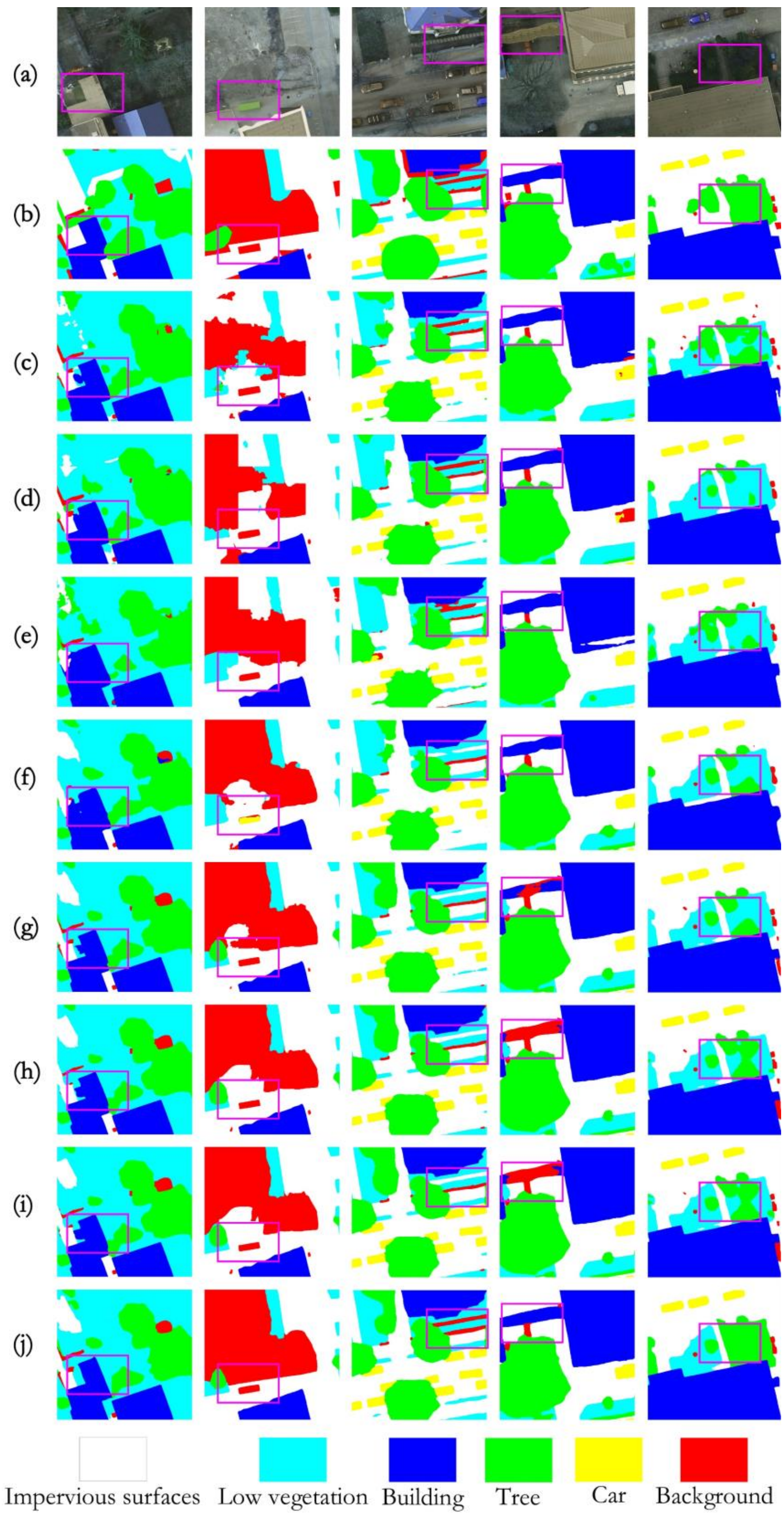

**Figure 6**. The visualization results of the comparative networks on the ISPRS Potsdam 2D test dataset: (a) the RGB image; (b) the truth labels; (c) PSPNet; (d) DeeplabV3+; (e) HRNet; (f) SwinUperNet; (g)HRFormer; (h) Segformer; (i) FT-UNetFormer; and (j) UDHF$^2$-Net. The purple boxes highlight the edge segmentation details.

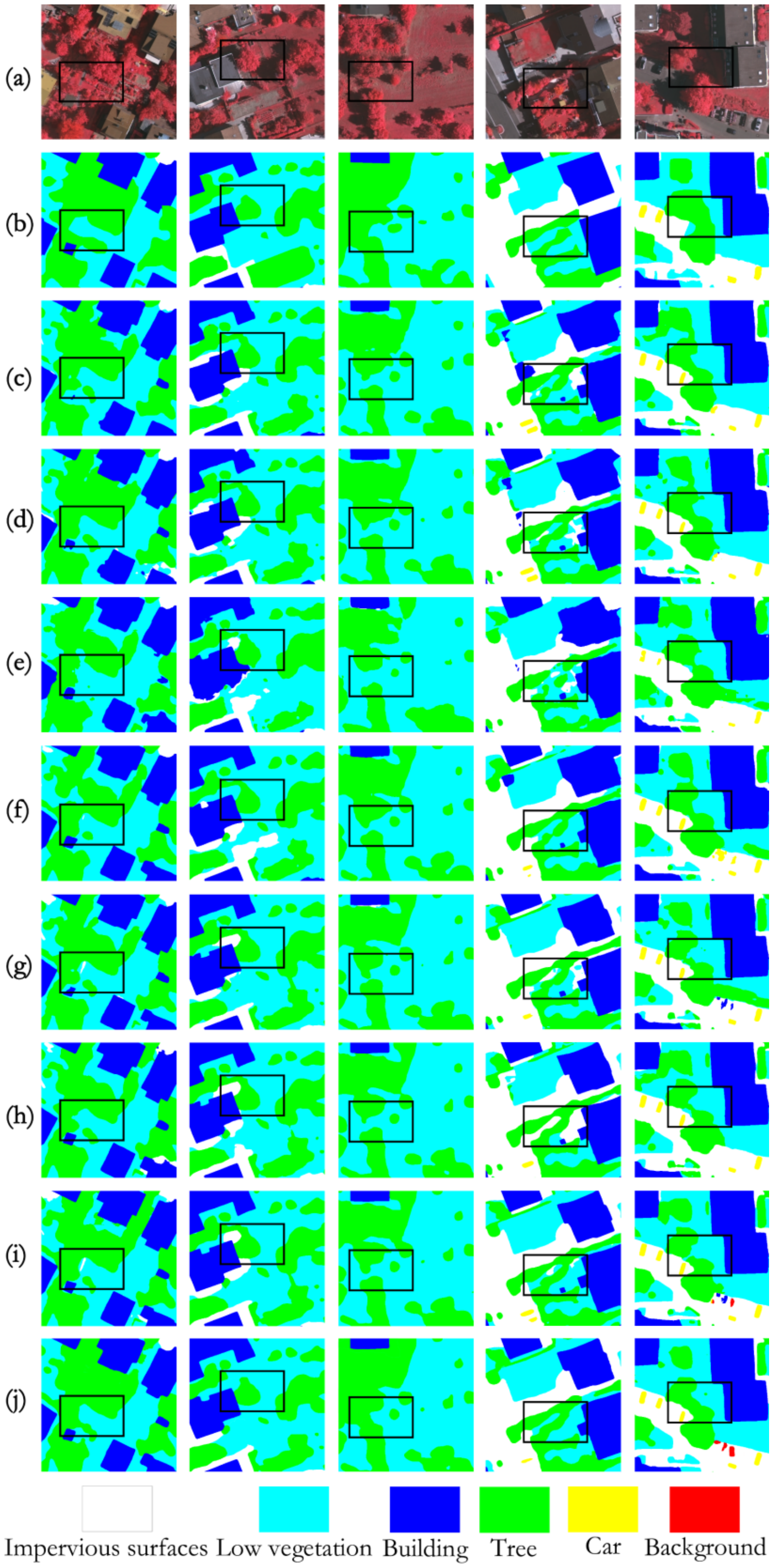

**Figure 7**. The visualization results of the comparative networks on the ISPRS Vaihingen 2D test dataset: (a) the RGB image; (b) the truth labels; (c) PSPNet; (d) DeeplabV3+; (e) HRNet; (f) SwinUperNet; (g)HRFormer; (h) Segformer; (i) FT-UNetFormer; and (j) UDHF$^2$-Net. The black boxes highlight the edge segmentation details.

Especially, for some classes of regular objects that are sensitive to edges such as buildings, impervious surfaces and cars, UDHF$^2$-Net achieves the highest values in F1 and IoU than all compared SOTA networks on all benchmark datasets. Additionally, the results of segmentation and visualization on Potsdam and Vaihingen dataset also illustrates the superiority and effectiveness of the proposed UDHF$^2$-Net as shown in Fig.6 and Fig.7.

**Table 3**. Quantitative analysis results on LoveDA test dataset for compared SOTA DNNs with mIoU (%). The highest values are highlighted in bold for every evaluation metric.

| Method | Backbone | Background | Building | Road | Water | Barren | Forest | Agriculture | mIoU |
|---|---|---|---|---|---|---|---|---|---|
| UNet | ResNet50 | 42.94 | 55.79 | 54.30 | 78.47 | 17.56 | 44.50 | 57.88 | 50.21 |
| UNet++ | ResNet50 | 43.20 | 55.73 | 55.36 | 78.82 | 18.32 | 44.49 | 57.56 | 50.50 |
| DeepLabv3+ | ResNet50 | 43.43 | 54.18 | 55.17 | 78.24 | 16.48 | 42.71 | 59.60 | 49.97 |
| PAN | ResNet50 | 43.99 | 55.22 | 53.66 | 78.12 | 16.35 | 44.50 | 59.86 | 50.24 |
| Semantic-FPN | ResNet50 | 43.43 | 53.77 | 55.79 | 78.76 | 19.62 | 44.37 | 59.87 | 50.80 |
| PSPNet | ResNet50 | 44.76 | 54.79 | 55.86 | 78.68 | 14.49 | 45.37 | 57.62 | 50.22 |
| LinkNet | ResNet50 | 43.40 | 54.70 | 54.81 | 78.25 | 17.27 | 44.63 | 57.08 | 50.02 |
| FarSeg | ResNet50 | 43.15 | 55.41 | 55.91 | 78.88 | 16.51 | 43.94 | 56.96 | 50.11 |
| FactSeg | ResNet50 | 42.51 | 54.62 | 55.88 | 77.96 | 16.51 | 44.72 | 57.81 | 50.00 |
| HRNet | HRNetV1 | 44.45 | 57.36 | 58.41 | 79.30 | 16.88 | 44.98 | 59.09 | 51.51 |
| Bi-FPN | ResNet50 | 43.98 | 55.99 | 55.42 | 78.92 | 18.24 | 43.85 | 59.94 | 50.91 |
| LoveNAS | ResNet50 | 45.39 | 58.86 | 59.45 | 79.39 | 13.76 | 43.94 | 65.64 | 52.34 |
| UNetFormer | ResNet18 | 44.70 | 58.80 | 54.90 | 79.60 | 20.10 | 46.00 | 62.50 | 52.40 |
| Segmenter | ViT-Tiny | 38.00 | 50.70 | 48.70 | 77.40 | 13.30 | 43.50 | 58.20 | 47.10 |
| SwinUperNet | Swin-Tiny | 43.30 | 54.30 | 54.30 | 78.70 | 14.90 | 45.30 | 59.60 | 50.00 |
| DC-Swin | Swin-Tiny | 41.30 | 54.50 | 56.20 | 78.10 | 14.50 | 47.20 | 62.40 | 50.60 |
| BANet | ResT-Lite | 43.70 | 51.50 | 51.10 | 76.90 | 16.60 | 44.90 | 62.50 | 49.60 |
| TransUNet | ViT-R50 | 43.00 | 56.10 | 53.70 | 78.00 | 9.30 | 44.90 | 56.90 | 48.90 |
| Ours | — | **47.34** | **61.93** | **60.59** | **81.13** | **21.55** | **48.41** | **65.95** | **55.27** |

*4.2.2 Ablation experiments of semantic segmentation*

This section presents detailed ablation experiments to evaluate the rationality and effectiveness of every components of the proposed UDHF$^2$-Net on Potsdam, Vaihingen, and LoveDA test datasets.

Table 4 validates that this proposed frequency-wise feature representation strategy is effective to objectively represent the spatial distribution patterns with coexistence of spatial stationary and non-stationary. To display the advantages of this proposed strategy, two single frequency-wise feature representation strategies are compared as following: UDHF$^2$-Net with the retained spatially frequency-wise stationary streams, and UDHF$^2$-Net with the retained spatially frequency-wise non-stationary streams. According to this result, it is observed that this proposed strategy outperforms the single frequency-wise feature representation in the metrics values of Mean F1, OA and mIoU on all chosen benchmark datasets.

Table 5 reports the effectiveness of the proposed MUDM. With the assistance of MUDM, the initial segmentation result is further optimized to improve the MeanF1 value by 1.16% and 1.23%, and the OA value by 1.85% and 2.25% on the Potsdam and Vaihingen datasets respectively. It is also help to increase the mIoU by 1.07%, 1.37% and 2.09% on the Potsdam, Vaihingen and LoveDA datasets respectively. In Fig.8, MUDM demonstrates the outstanding performing to recover the occluded building effectively.

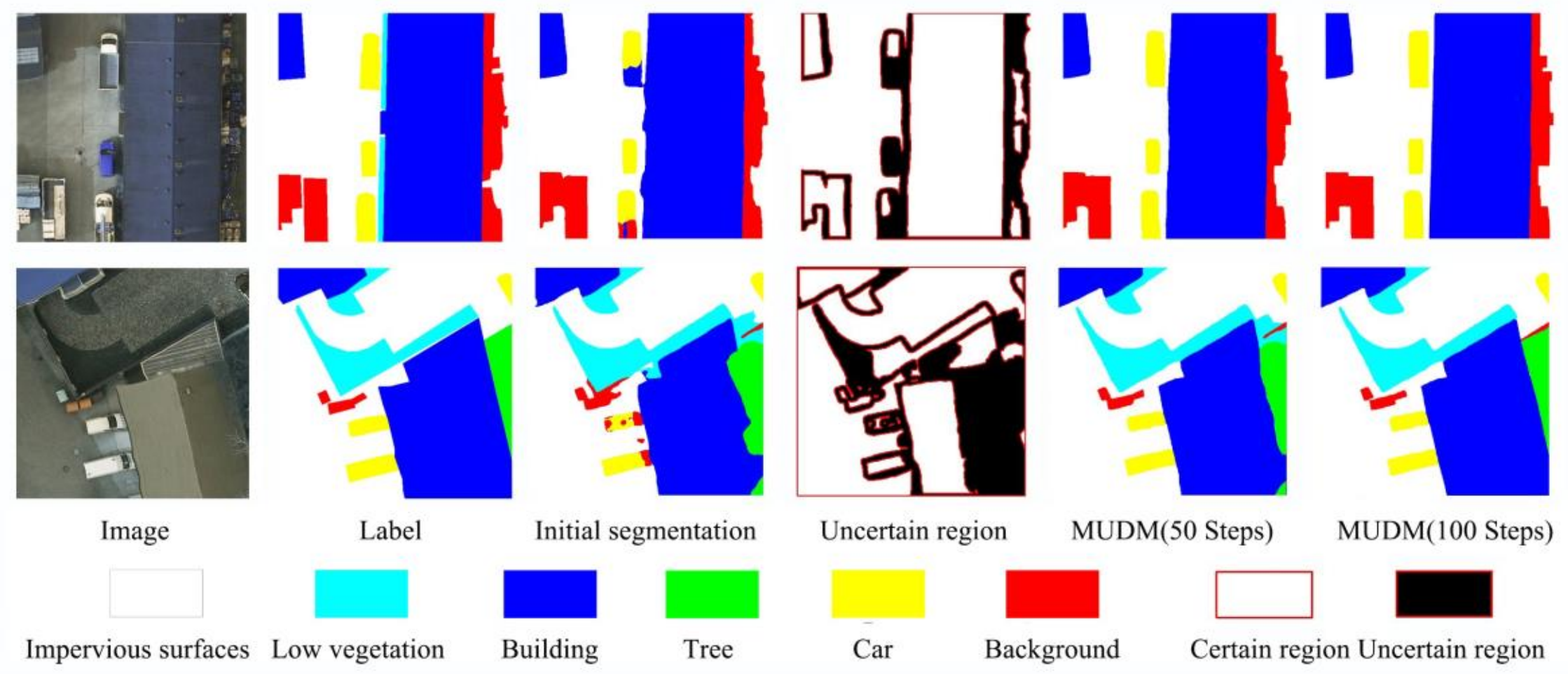

**Figure 8**. The visualization results of MUDM.

Table 6 indicates the significant performance of the proposed SHCP with MFD. Compared to other some outstanding decoders, i.e., UPerNet (Xiao et al., 2018) and OCR (Yuan et al., 2020), our proposed decoder achieves the improvement gains at least 1.99%, 2.18% in MeanF1, 2.18% and 2.32% in OA on the Potsdam and Vaihingen datasets respectively. Additionally, it also improves the mIoU values at least 3.24%, 2.27% and 1.40% on the Potsdam, Vaihingen and LoveDA datasets respectively. The above results reveal that this proposed decoder could effectively combine spatially frequency-wise stationary and non-stationary features to achieving better segmentation results compared with the current decoders.

**Table 4.** Ablation experiment of frequency-wise stationary and non-stationary for semantic segmentation. The highest values are highlighted in bold for every evaluation metric (%).

| Method | | UDHF$^2$-Net | | |
|---|---|---|---|---|
| Frequency-wise stationary | | √ | × | √ |
| Frequency-wise non-stationary | | × | √ | √ |
| ISPRS Potsdam 2D test dataset | MeanF1 | 94.08 | 94.21 | **95.78** |
| | OA | 93.97 | 94.01 | **95.42** |
| | mIoU | 90.82 | 91.06 | **92.00** |
| ISPRS Vaihingen 2D dataset | MeanF1 | 90.97 | 91.13 | **93.11** |
| | OA | 91.84 | 91.02 | **93.57** |
| | mIoU | 85.85 | 85.96 | **87.31** |
| LoveDA test dataset | mIoU | 53.98 | 54.07 | **55.27** |

**Table 5**. Ablation experiment of MUDM for semantic segmentation. The highest values are highlighted in bold for every evaluation metric (%).

| Method | | UDHF$^2$-Net | |
|---|---|---|---|
| SHCP | | √ | √ |
| MUDM | | × | √ |
| ISPRS Potsdam 2D test dataset | MeanF1 | 94.62 | **95.78** |
| | OA | 93.57 | **95.42** |
| | mIoU | 90.93 | **92.00** |
| ISPRS Vaihingen 2D dataset | MeanF1 | 91.88 | **93.11** |
| | OA | 91.32 | **93.57** |
| | mIoU | 85.94 | **87.31** |
| LoveDA test dataset | mIoU | 53.18 | **55.27** |

**Table 6**. Ablation experiment of SHCP with MFD for semantic segmentation. The highest values are highlighted in bold for every evaluation metric (%).

| Method | | UDHF$^2$-Net | | |
|---|---|---|---|---|
| Decoder | | UPerNet | OCR | SHCP with MFD |
| ISPRS Potsdam 2D test dataset | MeanF1 | 93.72 | 93.79 | **95.78** |
| | OA | 93.18 | 93.24 | **95.42** |
| | mIoU | 88.68 | 88.76 | **92.00** |
| ISPRS Vaihingen 2D dataset | MeanF1 | 90.76 | 90.93 | **93.11** |
| | OA | 91.12 | 91.25 | **93.57** |
| | mIoU | 84.96 | 85.04 | **87.31** |
| LoveDA test dataset | mIoU | 53.76 | 53.87 | **55.27** |

*4.3 Change detection results*

*4.3.1 Comparison experiments of change detection*

(1) Experimental results

**Table 7**. Quantitative analysis results on WB-CD test dataset for compared SOTA DNNs. The highest values are highlighted in bold for every evaluation metric (%). (Some values are null because they were not provided in the original paper.)

| Method | Precision | Recall | F1 score | IoU |
|---|---|---|---|---|
| AMTNet | 92.86 | 91.99 | 92.42 | 85.64 |
| FC-EF | 80.87 | 75.43 | 78.05 | 64.01 |
| FC-Siam-conc | 68.62 | 87.30 | 76.84 | 62.39 |
| FC-Sima-diff | 70.45 | 77.62 | 73.86 | 58.56 |
| DSIFN/IFN | 91.47 | 81.57 | 86.24 | 75.99 |
| DTCDSCN | — | 89.32 | 89.01 | 79.08 |
| SNUNet | 83.25 | 91.35 | 87.11 | 77.17 |
| BiT | 86.64 | 81.48 | 83.98 | 72.39 |
| ChangeFormer | 92.89 | 85.60 | 88.82 | 79.89 |
| MTCNet | — | 91.90 | 82.65 | 70.43 |
| SGSLN/512 | 96.11 | 93.64 | 94.86 | 90.22 |
| W-Net | 94.76 | 88.88 | 91.72 | — |
| P2V-CD | 95.48 | 89.47 | 92.38 | — |
| ChangeCLIP | 96.02 | 93.58 | 94.78 | 90.08 |
| Ours | **97.40** | **95.49** | **96.44** | **93.12** |

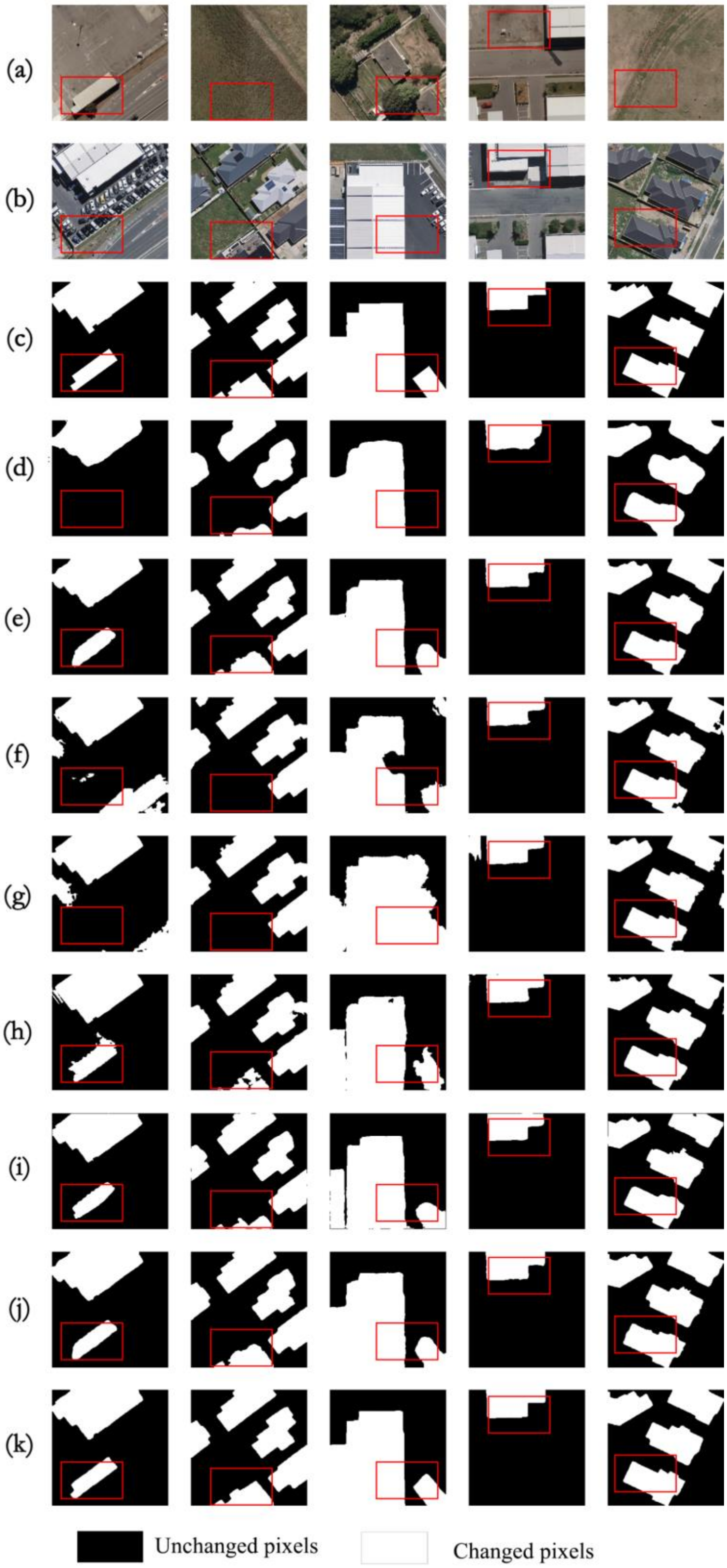

**Figure 9**. The visualization results of the different networks on the WB-CD dataset: (a) the RGB image of the first date; (b) the RGB image of the second date; (c) the truth labels; (d) FC-Siam-conc; (e) FC-Siam-diff; (f) SNUNet; (g) BiT; (h) IFNet; (i) P2V-CD; (j) ChangeFormer; and (k) semi-pseudo-Siamese UDHF$^2$-Net. The red boxes highlight the edge extraction details.

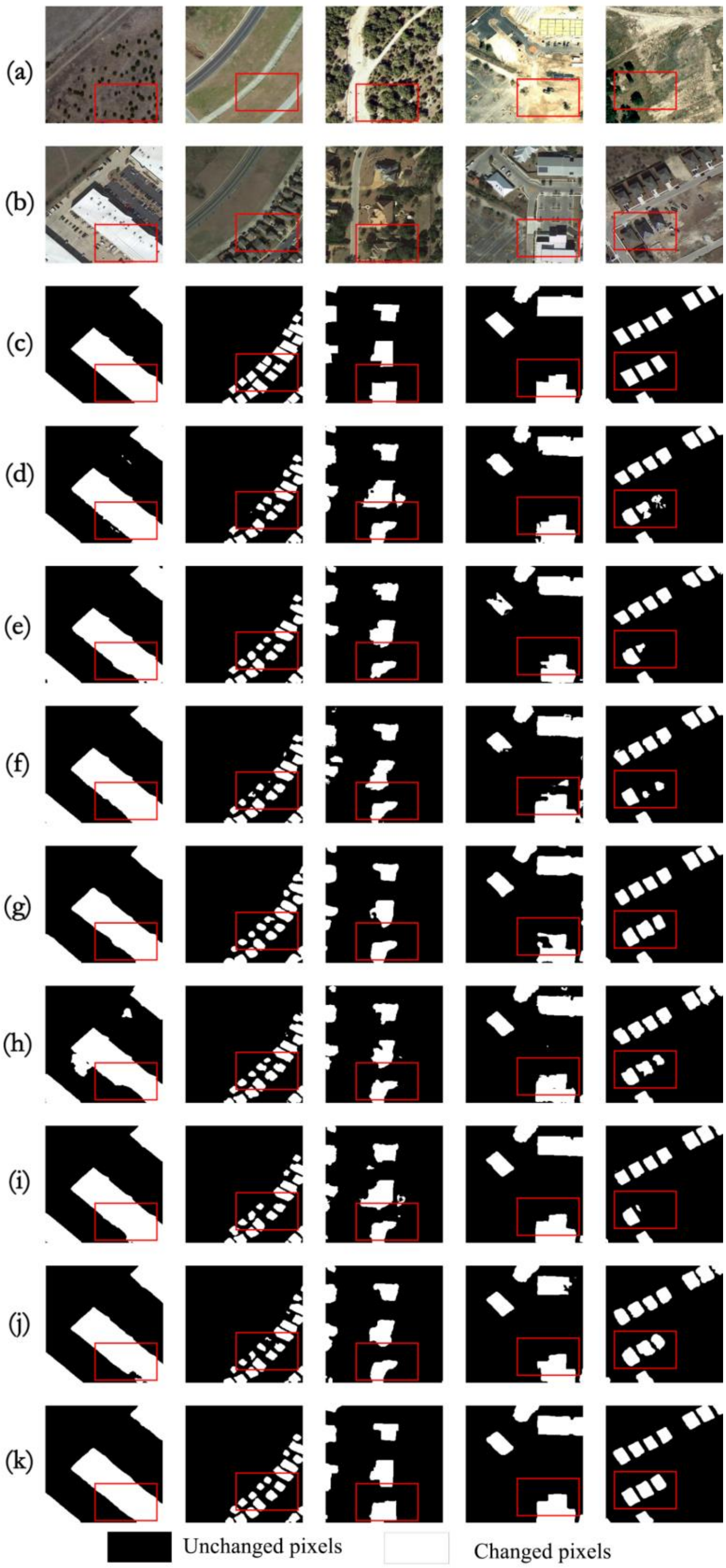

**Figure 10**. The visualization results of the different networks on the LEVIR-CD dataset: (a) the RGB image of the first date; (b) the RGB image of the second date; (c) the truth labels; (d) FC-Siam-conc; (e) FC-Siam-diff; (f) SNUNet; (g) BiT; (h) IFNet; (i) P2V-CD; (j) ChangeFormer; and (k) semi-pseudo-Siamese UDHF$^2$-Net. The red boxes highlight the edge extraction details.

**Table 8**. Quantitative analysis results on the LEVIR-CD test dataset for compared SOTA DNNs. The highest values are highlighted in bold for every evaluation metric (%). (Some values are null because they were not provided in the original paper.)

| Method | Precision | Recall | F1 score | IoU |
|---|---|---|---|---|
| AMTNet | 91.82 | 89.71 | 90.76 | 83.08 |
| FC-EF | 86.91 | 80.17 | 83.4 | 71.53 |
| FC-Siam-conc | 91.99 | 76.77 | 83.69 | 71.96 |
| FC-Sima-diff | 89.53 | 83.31 | 86.31 | 75.92 |
| DSIFN/IFN | 88.53 | 86.83 | 87.67 | 78.05 |
| SNUNet | 89.18 | 87.17 | 88.16 | 78.83 |
| BiT | 89.24 | 89.37 | 89.31 | 80.68 |
| ChangeFormer | 92.05 | 88.80 | 90.40 | 82.48 |
| MTCNet | — | 89.62 | 90.24 | 82.22 |
| GAS-Net | 91.82 | 90.62 | 91.21 | — |
| W-Net | 91.24 | 89.21 | 90.31 | — |
| SGSLN/512 | 93.07 | **91.61** | 92.33 | 85.76 |
| P2V-CD | 93.32 | 90.60 | 91.94 | — |
| ChangerEX | 93.61 | 90.56 | 92.06 | — |
| CMID | — | — | 91.72 | — |
| GFM | — | — | 91.73 | — |
| SkySense | — | — | 92.58 | — |
| MSP-CD | 86.08 | 93.93 | 89.83 | — |
| ChangeCLIP | 93.40 | 90.67 | 92.01 | 85.20 |
| Ours | **96.03** | **93.99** | **95.00** | **90.47** |

**Table 9**. Quantitative analysis results on the SYSU-CD test dataset for compared SOTA DNNs. The highest values are highlighted in bold for every evaluation metric.

| Method | Precision | Recall | F1 | IoU |
|---|---|---|---|---|
| DSAMNet | 74.81 | 81.86 | 78.18 | 64.18 |
| CDNet | 79.34 | 77.29 | 78.3 | 64.34 |
| ISNet | 76.41 | 80.27 | 78.29 | 64.44 |
| SNUNet | 83.58 | 75.87 | 79.54 | 66.02 |
| L-UNet | 81.24 | 78.08 | 79.63 | 66.15 |
| IFN | 80.98 | 79.37 | 80.17 | 66.9 |
| DPCC-Net | 81.05 | 78.93 | 79.97 | 66.63 |
| DARNet | 83.04 | 79.11 | 81.03 | 68.1 |
| ICIFNet | 83.37 | 78.51 | 80.74 | 68.12 |
| SSANet | 82.48 | 79.73 | 81.08 | 68.18 |
| AFCF3D-Net | 82.3 | 83.88 | 83.11 | 71.09 |
| ChangeCLIP | 87.16 | 79.80 | 83.32 | 71.41 |
| MCHA-Net | 75.86 | 69.00 | 68.16 | 58.31 |
| Our | **90.38** | **85.22** | **87.72** | **78.13** |

As can be seen in Table 7, 8, and 9, semi-pseudo-Siamese UDHF$^2$-Net proves the advancement compared with the existing SOTA DNNs in change detection task, which achieved the highest values of all evaluation metrics on all selected benchmark datasets. The detailed reports are shown as following: the proposed network achieved the superior Precision values of 97.40%, 96.03% and 90.38%, the higher Recall values of 95.49%, 93.99% and 85.22%, the bigger F1 values of 96.44%, 95.00% and 87.72%, as well as impressive IoU values of 93.12%, 90.47% and 78.13% on the test datasets of WHU-CD, LEVIR-CD and SYSU-CD, respectively. Concretely, the visual representation on WHU-CD and LEVIR-CD datasets is provided to verify illustrates the representation of the edge segmentation details compared with optimal SOTA network as shown in Fig.9 and Fig.10.

*4.3.3 Ablation experiments of change detection*

Detailed ablation experiments were conducted to evaluate the rationality and effectiveness of every components of the proposed semi-pseudo-Siamese UDHF$^2$-Net on WB-CD, LEVIR-CD and SYSU-CD test dataset.

**Table 10**. Ablation experiment of frequency-wise stationary and non-stationary for change detection. The highest values are highlighted in bold for every evaluation metric (%).

| Method | | Semi-Pseudo-Siamese UDHF$^2$-Net | | |
|---|---|---|---|---|
| Frequency-wise stationary | | √ | × | √ |
| Frequency-wise non-stationary | | × | √ | √ |
| LEVIR-CD test dataset | Precision | 94.47 | 94.62 | **96.03** |
| | Recall | 92.44 | 92.59 | **93.99** |
| | F1 | 93.33 | 93.49 | **95.00** |
| | IoU | 88.93 | 89.23 | **90.47** |
| WB-CD test dataset | Precision | 96.05 | 96.21 | **97.40** |
| | Recall | 93.89 | 94.01 | **95.49** |
| | F1 score | 94.96 | 95.10 | **96.44** |
| | IoU | 91.76 | 91.83 | **93.12** |
| SYSU-CD test dataset | Precision | 88.11 | 88.23 | **90.38** |
| | Recall | 82.79 | 82.97 | **85.22** |
| | F1 | 85.37 | 85.52 | **87.72** |
| | IoU | 75.74 | 75.86 | **78.13** |

**Table 11**. Ablation experiment of MUDM for change detection. The highest values are highlighted in bold for every evaluation metric (%).

| Method | | Semi-Pseudo-Siamese UDHF$^2$-Net | |
|---|---|---|---|
| MUDM | | × | √ |
| LEVIR-CD test | Precision | 95.01 | **96.03** |
| dataset | Recall | 92.84 | **93.99** |
| | F1 | 93.91 | **95.00** |
| | IoU | 89.24 | **90.47** |
| WB-CD test dataset | Precision | 96.22 | **97.40** |
| | Recall | 92.05 | **95.49** |
| | F1 score | 94.09 | **96.44** |
| | IoU | 90.86 | **93.12** |
| SYSU-CD test dataset | Precision | 88.84 | **90.38** |
| | Recall | 83.07 | **85.22** |
| | F1 | 85.96 | **87.72** |
| | IoU | 76.08 | **78.13** |

**Table 12**. Ablation experiment of differencing and Siamese architecture for change detection. The highest values are highlighted in bold for every evaluation metric (%).

| Method | | SHCP-based differencing UDHF$^2$-Net | Semi-Pseudo-Siamese UDHF$^2$-Net |
|---|---|---|---|
| LEVIR-CD test dataset | Precision | 94.25 | **96.03** |
| | Recall | 91.47 | **93.99** |
| | F1 | 92.82 | **95.00** |
| | IoU | 88.12 | **90.47** |
| WB-CD test dataset | Precision | 94.14 | **97.4** |
| | Recall | 90.93 | **95.49** |
| | F1 score | 92.51 | **96.44** |
| | IoU | 89.24 | **93.12** |
| SYSU-CD test dataset | Precision | 87.62 | **90.38** |
| | Recall | 82.16 | **85.22** |
| | F1 | 84.80 | **87.72** |
| | IoU | 74.10 | **78.13** |

Table 10 indicates the effectiveness of the proposed coexistence strategy of spatially frequency-wise stationary and non-stationary features to objectively understand the remote sensing senses compared with two single frequency-wise feature representation strategies only including spatially stationary and non-stationary features. The results shown that this proposed strategy demonstrates the better performance gains at least by 1.24% of IoU, 1.34% of F1, 1.19% of Precision and 1.48% of Recall on the three chosen benchmark datasets. That indicates that the complementarity of multiple stationary and non-stationary feature streams could effectively improve the accuracy in change detection task.

Table 11 reports the superiority of the proposed MUDM. The results demonstrate that MUDM also has the advantages to improve the accuracy of change detection. It could exceed the network without MUDM up to the greater values by 2.26% of IoU on WHU-CD dataset, 2.35% of F1 on WHU-CD dataset, 1.54% of Precision on SYSU-CD dataset and 3.34% of Recall on WHU-CD dataset. And the visualization results of MUDM are shown in Fig.11.

Table 12 reports the effectiveness of semi-pseudo-Siamese architecture for the remote sensing change detection task. The SHCP-based difference-based architecture is compared with to achieve change intensity information. Compared with the difference-based architecture, the semi-pseudo-Siamese UDHF$^2$-Net obtains the better improvement by 2.35%, 3.88% and 4.03% of IoU, 2.18%, 3.93% and 2.92%of F1, 1.78%, 3.26% and 2.76 of Precision, and 2.52%, 4.56% and 3.06 of Recall on LEVIR-CD, WHU-CD and SYSU-CD dataset respectively.

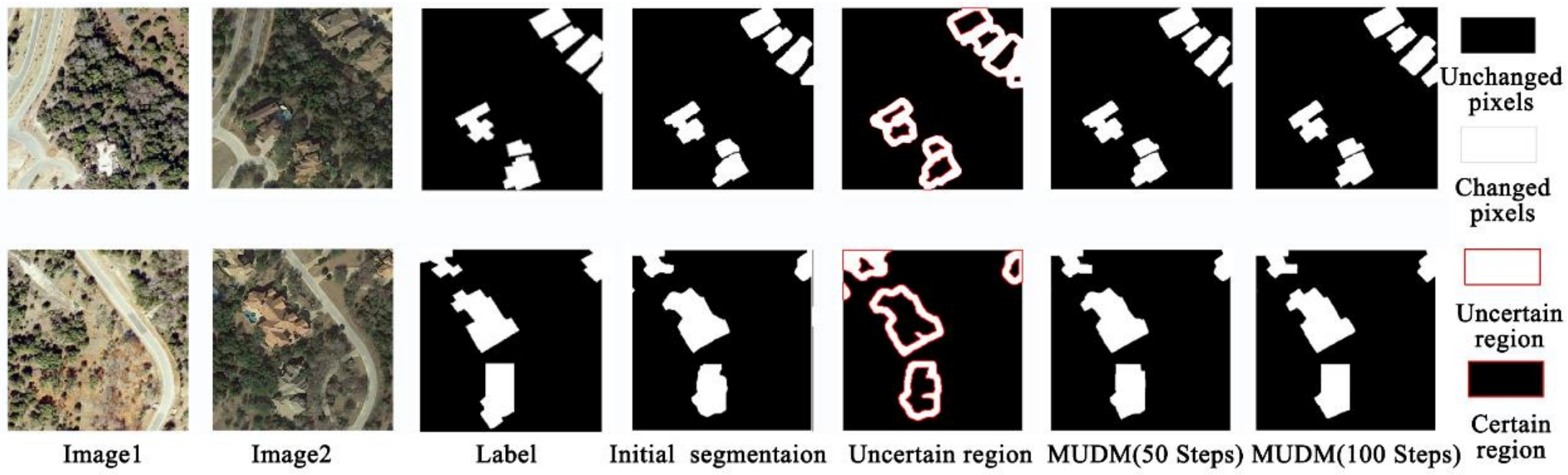

**Figure 11**. The visualization results of MUDM in change detection.

*4.4. Discussion*

*4.4.1 The accuracy impact by frequency representation*

The proposed SHCP could effectively improve segmentation accuracy by the objective representation of the spatial distribution patterns with the coexistence of spatial stationary and non-stationary. As demonstrated in Table 4 and 10, the ablation experimental results showcase the superiority of this coexisted frequency-wise feature representation scheme over the single feature representation scheme (i.e., using only stationarity or non-stationarity features) whether in both semantic segmentation and change detection tasks. The rationality of this method is attributed to the intrinsic mechanisms that it has attempted and resolved the complementarity between spatial frequency-wise stationarity and non-stationarity features. It is similar to combining the ability of global feature representation and local feature representation to objective represent of the spatial distribution patterns of remote sensing scenes, which has been validated as an effective approach (Jia and Yao, 2023). Different from the current researches, we further have explored to enhance their frequency representation of capturing global consistent information by spatially frequency-wise stationary features and local abrupt change information by spatially frequency-wise non-stationary features. Thus, that is conducive to improving the segmentation accuracy of RSII.

*4.4.2 The improvement of edge extraction accuracy*

The proposed UDHF$^2$-Net demonstrates the superiority of edge extraction. As shown in Table 1, 2, 3, 7, 8 and 9, UDHF$^2$-Net achieved excellent performance than the other compared SOTA networks. The scientificity of UDHF$^2$-Net benefits from the following aspects:

(1) High frequency information can improve the edge extraction accuracy. Importantly, high frequency features are sensitive for edge details (Shan et al., 2021). The proposed SHCP could deliver high-frequency feature stream in the whole encoder-decoder process to remain high-fidelity edge information and reduce the impact of downsampling. And HD-Net (Li et al., 2024) also have verified that the architecture similar to HRFormer is help to achieve lossless edge performance. In addition, their respective advantages of the aforementioned spatially stationary and non-stationary high-frequency features can be complemented, which enhances the location and extraction ability of edge details.

(2) Edge loss function can optimize the edge extraction accuracy. To tackle the issue of edge detail degradation, numerous researches have placed special emphasis on edge loss function to improves edge discrimination (Zheng et al., 2020). Furthermore, we introduced the uncertain loss function to optimize the edge with the constraint of prior probability, which is an effective way to improve the edge extraction accuracy.

(3) Uncertainty diffusion model that is a self-supervised learning strategy can reduce the noises (e.g., mixed pixel, edge occlusion and registration error etc.) and recovering the truth labels. As shown in Table 5 and 11, the experimental results illustrate that the proposed MUDM could further optimize the initial results, which could gradually reduce the multiple radiometric noises generated by the geo-knowledge-based diffusion models. Current researches about denoising diffusion probabilistic model has achieved excellent performance in repairing occlusion issue such as removing cloud (Zou et al., 2024). In addition, this work could better distinguish mixed pixels, recover the buildings or roads occluded by vegetation and reduce the visual errors. This way could further optimize the edge extraction effect of initial segmentation result.

*4.4.3 The improvement of false change detection*

The proposed semi-pseudo-Siamese UDHF$^2$-Net could adaptively address false detection caused by geometric registration error with self-supervised learning strategy in change detection task. As shown in Table 7, 8 and 9, the proposed network achieves the best F1 and IoU values on WB-CD, LEVIR-CD and SYSU-CD test datasets compared with the current SOTA networks. The semi-pseudo-Siamese UDHF$^2$-Net is a variant of UDHF$^2$-Net, which has the aforementioned advancements shown in section 4.4.1 and 4.4.2. Besides, it still has the following superiorities:

Firstly, semi-pseudo-Siamese architecture is a trade-off way to balance the computational complexity and the accuracy of change detection. It freezes the majority of network layers to reduce the computational complexity, and remains some layers to be fine-tuned for effectively adapting their respective characteristics. That's could help the network to increase the ability of identify difference features for improving the accuracy of change detection.

Secondly, the anti-false-change ability for geometric registration error is improved to extract better change results, which benefits by the following aspects: (1) The generated negative samples with simulated registration error are trained to enhance the robustness of the proposed network; (2) The generated noises (e.g., the aforementioned noises and registration error) in uncertain edge are gradually reduced by the embedded MUDM, which is useful for refining edge continuity and improving the accuracy of change detection.

## 5. Conclusion

To performance highly-accurate planetary observation, UDHF$^2$-Net is the first to be proposed to systematically handling each step of RSII in both semantic segmentation and change detection tasks. This UDHF$^2$-Net could draw the following conclusions: (1) the spatially-stationary-and-non-stationary high-frequency connection paradigm (SHCP) could effectively improve the accuracy of RSII by fusing the spatially frequency-wise stationary and non-stationary features to objectively represent the spatial distribution patterns; (2) the mask-and-geo-knowledge-based uncertainty diffusion module (MUDM) could anti-false-change ability and improve the accuracy of edge extraction; (3) the semi-pseudo-Siamese UDHF$^2$-Net could improve the accuracy of change detection.

In the future, we will further explore the research about frequency-wise remote sensing foundation model. Meanwhile, we hope that the proposed work could encourages more scholars dedicated to deeply explore the potential and possibility of interpretable frequency-wise machine learning.

## Acknowledgements

The authors are grateful for the comments and contributions of the editors, anonymous reviewers and the members of the editorial team. This work was supported by the Key Program of the National Natural Science Foundation of China under Grant Nos. 42030102, National Natural Science Foundation of China (NSFC) under Grant Nos. 41771493 and 41101407, and the Fundamental Research Funds for the Central Universities under Grant CCNU22QN019.

## Declaration of competing interest

The authors declare that they have no known competing financial interests or personal relationships that could have appeared to influence the work reported in this paper.